\documentclass[journal]{IEEEtran}
\usepackage{amsmath,amsfonts}
\usepackage{algorithmic}
\usepackage{array}
\usepackage[caption=false,font=normalsize,labelfont=sf,textfont=sf]{subfig}
\usepackage{textcomp}
\usepackage{stfloats}
\usepackage{url}
\usepackage{verbatim}
\usepackage{graphicx}

\usepackage{romannum}
\usepackage{amssymb}
\usepackage{diagbox}
\usepackage{makecell}
\usepackage{amssymb}
\usepackage{bm}
\usepackage{cite}
\usepackage{mathtools}
\usepackage{booktabs}
\usepackage{color}
\usepackage{soul}

\def\BibTeX{{\rm B\kern-.05em{\sc i\kern-.025em b}\kern-.08em
    T\kern-.1667em\lower.7ex\hbox{E}\kern-.125emX}}
\usepackage{balance}
\begin{document}
\title{Virtual Elastic Tether: a New Approach for Multi-agent Navigation in Confined Aquatic Environments}
\author{Kanzhong Yao, Xueliang Cheng, Keir Groves, Barry Lennox, Ognjen Marjanovic, and Simon Watson
\thanks{Manuscript received xxx}
\thanks{This work was supported Chinese Scholarship Council-University of Manchester joint programme. The authors acknowledge
the support provided by EPSRC (Hot Robotics: EP/T011491/1)
and the Robotics and AI Collaboration (RAICo).}
\thanks{All authors are with Manchester Centre for Robotics and AI, Department of Electrical and Electronic Engineering, University of Manchester, UK. Please
direct correspondence to {\tt\small kanzhong.yao@postgrad.manchester.ac.uk}}%
}

\maketitle

\begin{abstract}
Underwater navigation is a challenging area in the field of mobile robotics due to inherent constraints in self-localisation and communication in underwater environments. Some of these challenges can be mitigated by using collaborative multi-agent teams. However, when applied underwater, the robustness of traditional multi-agent collaborative control approaches is highly limited due to the unavailability of reliable measurements. 
In this paper, the concept of a Virtual Elastic Tether (VET) is introduced in the context of incomplete state measurements, which represents an innovative approach to underwater navigation in confined spaces.
The concept of VET is formulated and validated using the Cooperative Aquatic Vehicle Exploration System (CAVES), which is a sim-to-real multi-agent aquatic robotic platform.
Within this framework, a vision-based Autonomous Underwater Vehicle-Autonomous Surface Vehicle leader-follower formulation is developed. Experiments were conducted in both simulation and on a physical platform, benchmarked against a traditional Image-Based Visual Servoing approach.
{Results indicate that the formation of the baseline approach fails under discrete disturbances,  when induced distances between the robots exceeds 0.6\,m in simulation and 0.3\,m in the real world. In contrast, the VET-enhanced system recovers to pre-perturbation distances within 5 seconds. Furthermore, results illustrate the successful navigation of VET-enhanced CAVES in a confined water pond where the baseline approach fails to perform adequately.}

\end{abstract}

\begin{IEEEkeywords}
underwater navigation,  visual servoing, multi-agent control, sim-to-real, extreme environment
\end{IEEEkeywords}

\section{INTRODUCTION}


\IEEEPARstart{C}{onfined} underwater environments are prevalent in various industries \cite{legacy_pond} such as liquid storage tanks, submerged tunnels, underwater pipelines, boatyards, and aging nuclear waste storage ponds \cite{fuel_pond}, where direct human access is challenging and hazardous.
Traditional methods such as divers, tethered Remote Operated Vehicles (ROVs), or stationary sensors either pose safety risks or fail to provide the {coverage} needed for effective monitoring.
Furthermore, these locations often require long-term monitoring, inspection, and maintenance to ensure environmental safety, operational integrity, and regulatory compliance \cite{Watson2020}.
 As a result, autonomous and semi-autonomous robotic systems are increasingly becoming indispensable tools for safe and efficient operation in these confined aquatic environments \cite{Zhang2023}. 
As these infrastructures age, undergo maintenance, or adapt to new regulatory requirements, the need for more robust and versatile robotic solutions for thorough characterization and monitoring becomes even more critical  \cite{oikawa2016r}.

\subsection{Motivation}
  \begin{figure}[tpb]
      \centering
       \includegraphics[width=0.48\textwidth]{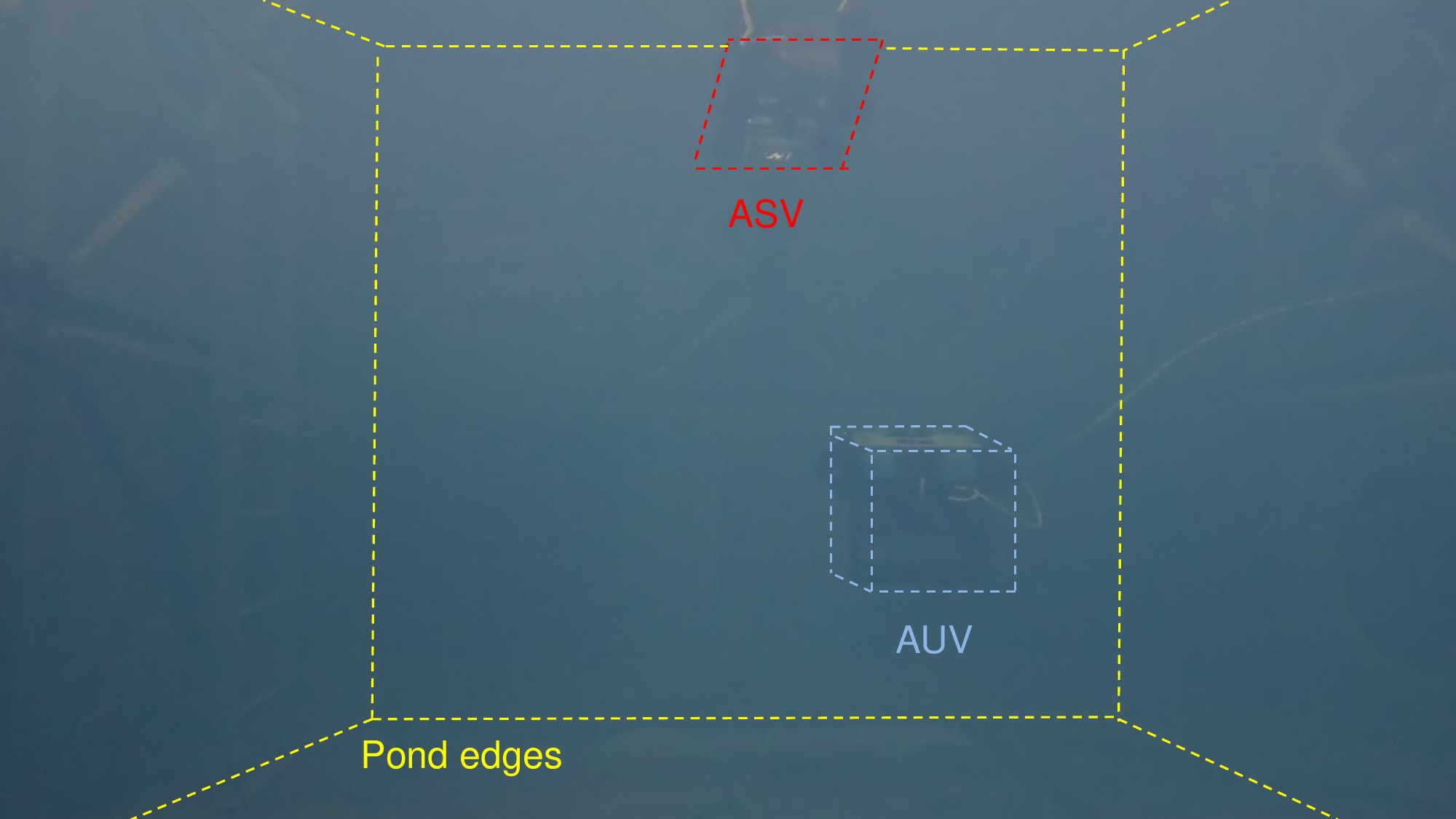}
      \caption{Example of confined underwater space with low visibility, where environmental features such as walls are extremely hard to extract, complicating the application of traditional methods such as SLAM or feature-based odometry. Note that the picture was taken by an underwater motion capture camera with 5.3K resolution.
      }
      \label{fig:caves_display1}
  \end{figure}
With advancements in control, localisation, and perception, robotic systems are increasingly deployed in diverse environments for exploratory \cite{Zhou2022} and inspection missions \cite{Bird2021,Jiang2022,GonzalezGarcia2020}. However, when considering applications in complex and confined aquatic environments, additional challenges arise that demand specialized consideration. Unlike air, ground-based, and open-ocean scenarios, these environments present unique challenges and conditions.

In order to enable robotic systems to explore, inspect and characterise such challenging confined aquatic environments, they must address the following environmental characteristics

\begin{itemize}
\item \textbf{Communication and self-localisation limitations:} Traditional methods of communication and localization, such as RF (Radio Frequency) and GPS, are ineffective underwater due to the aquatic conditions \cite{Qu2016}. In certain environments, especially those that are contaminated such as nuclear sites, robots are designed to be disposable. Given this context, the high costs associated with underwater optical communication modems make them an impractical choice.Additionally, poor lighting conditions underwater limit visibility (Fig.~\ref{fig:caves_display1}), complicating the application of traditional methods such as visual SLAM \cite{Lategahn2011}.

\item \textbf{Lack of external facilities:} Installation of auxiliary facilities, such as communication stations or localisation beacons \cite{Zhang2023}, are unavailable \cite{Watson2020} in many confined or restricted access environments \cite{legacy_pond}.
\item \textbf{Limited space:} The nature of confined aquatic environments restricts the maneuverability of robots \cite{Duecker_2020}. Therefore, high levels of autonomy and tetherless operation are often required for these robots.
\end{itemize}

While current underwater robots rely either on acoustics for positioning and navigation \cite{Zhang2023}, which is unsuitable in confined
spaces because of echo problems \cite{Yang_2017}, or the presence of dense and continuous features \cite{Duecker2020b} for self-localisation, which may not be possible in confined water ponds due to visibility constraints, the development and integration of novel navigation methods become imperative. 

One potential solution to the challenges highlighted above is the deployment of multiple collaborative robots. In such a multi-agent system, despite the visibility constraints of the environmental features, each robot can serve as a source of controllable visual features for its peers, thereby facilitating effective visual-based navigation and task execution. However, implementing such multi-agent systems introduces additional challenges. Specifically, ensuring robust and autonomous cooperative behavior becomes a critical issue, especially when communication is constrained and reliance on external infrastructure is not feasible.

  \begin{figure}[tpb]
      \centering
       \includegraphics[width=0.49\textwidth]{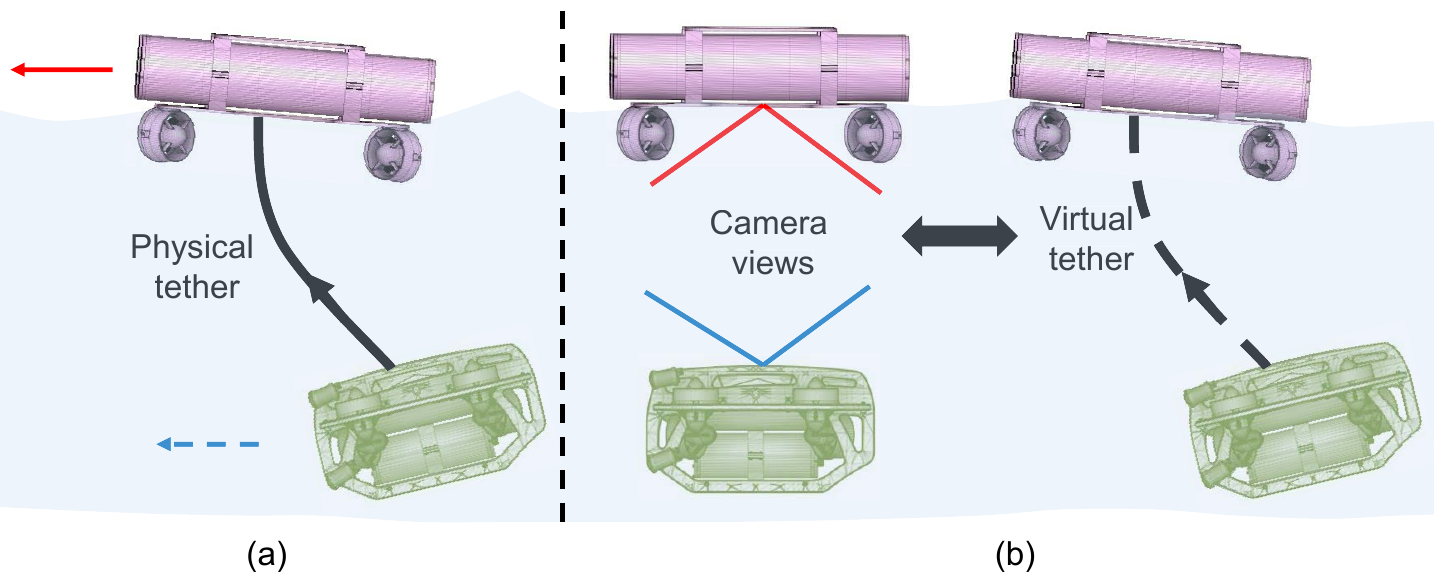}
      \caption{Illustration of multiple cooperative aquatic robots. a) displays a robust connection between the AUV (green) and ASV (pink) was achieved by facilitating a physical tether between the robots; b) shows the AUV and ASV are using direct LoS camera views to formulate a virtual connection.
      }
      \label{fig:vet_display}
  \end{figure}

To address this technical challenge, a new approach, referred to as \textbf{V}irtual \textbf{E}lastic \textbf{T}ether (VET),  is proposed in this paper for robust multi-agent navigation in confined aquatic environments.
VET simulates an imaginary tether to connect multiple robots using direct Line-of-Sight (LoS) camera views, allowing them to execute their individual maneuvers while maintaining a certain level of mutual proximity. Fig.~\ref{fig:vet_display} shows an example of VET on an Autonomous Underwater Vehicle (AUV) and Autonomous Surface Vehicle (ASV). 
While the term ``virtual tether'' has been employed in prior studies, such as that reported in \cite{Tolba2017}, the usage of ``virtual tether'' in these applications is limited to defining a search space based on underwater odometry, which is not pertinent to cooperative navigation.
Similarly, the virtual spring damper method utilised in \cite{Wiech2018} focuses on leader-follower formation control by designating specific roles to robots. However, this method does not account for the interactions between these leaders and followers.
As such, their approach closely mirrors the method considered in \cite{Yao2023}, which has demonstrated the instability of such configurations in underwater deployment.

Furthermore, to validate the effectiveness of VET and further explore its extended applications, we propose the \textbf{C}ooperative \textbf{A}quatic \textbf{V}ehicle \textbf{E}xploration \textbf{S}ystem (CAVES), a sim-to-real multi-agent aquatic robotic platform.  A more detailed description of CAVES in terms of the hardware and software packages is provided in Section \MakeUppercase{\romannumeral 6} and Section \MakeUppercase{\romannumeral 7}.

 
\subsection{Contribution}

The contributions of the approach presented in this paper can be summarised as follows: 

\begin{itemize}
     \item A new approach for multi-agent navigation in confined aquatic environments is proposed, which has been formulated in a generalised form. This approach can also be readily adapted for applications in other environments, such as communication and GPS denied indoor warehouses, featureless playgrounds, or even outer space.
     \item With the proposed multi-agent navigation approach, a VET-enhanced implementation of Image Based Visual Servoing (IBVS) leader-follower control is proposed. It is worth noting that, when compared to \cite{Wang2023,Lin2021,Bechlioulis2019a}, our implementation of IBVS is neither camera mode sensitive nor does it require camera calibration. Additionally, it demonstrates significant robustness improvements when deployed underwater, compared to \cite{Yao2023}. {Sim-to-real experimental validation was also conducted.}
\end{itemize}

To promote further exploration and development within this domain, the current CAVES simulation and vision-based formation control package have been made accessible as open-source resources\footnote{\url{git@github.com:drunkbot/CAVES-simulation.git}} for the community. 

To the best of the authors’ knowledge, CAVES, with the implementation of VET, is the first multi-agent underwater robotic platform that is capable of conducting robust cooperative navigation in communication denied confined aquatic environments. It is important to note that, this is achieved even under conditions of low underwater visibility for which conventional methods, such as Qualisys Motion Capture Systems\footnote{\url{https://www.qualisys.com/}}, have been shown not to be able to provide the trajectory ground truth in a 3\,m$\times$4\,m water tank\footnote{\url{https://hotrobotics.co.uk/equipment/simulated-fuel-pond/}}.

The rest of this article is organised as follows. Section \MakeUppercase{\romannumeral 2} reviews related work focused on addressing the challenge of underwater navigation. Section \MakeUppercase{\romannumeral 3} defines the navigation problem as a multi-agent leader-follower scenario. A generalised formulation of VET is presented in Section \MakeUppercase{\romannumeral 4} and then instantiated to a real-world applicable form for CAVES in Section \MakeUppercase{\romannumeral 5}. Section \MakeUppercase{\romannumeral 6} and Section \MakeUppercase{\romannumeral 7} demonstrate the superior performance of the sim-to-real application of VET by benchmarking it against a traditional IBVS approach. Section \MakeUppercase{\romannumeral 8} concludes this article and outlines potential directions for future extensions of VET. 
\section{Related Work}
Literature related to the work presented in this paper is briefly reviewed in the following. Proceeding subsections cover existing underwater robotic platforms, collective robot motion techniques, and visual servoing multi-agent navigation in underwater environments, which define the problem that this work aims to address. 

\subsection{Underwater Robots for Inspection and Exploration}
In recent years, various robotic systems have been developed to safely and efficiently explore confined underwater environments. Examples of such systems include AVEXIS \cite{Griffiths_2016}, the HippoCampus AUV \cite{Duecker2020b}, the SAM AUV \cite{Bhat2020}, and the LoCO AUV \cite{Edge2020a}. Nonetheless, these platforms often have limitations: they may depend on a tether for communication with a ground station PC, necessitate pre-installed auxiliary navigational infrastructure, or be restricted to surface-level operations. The manoeuvrability of tethered robots in confined spaces is highly limited, and there is a risk of the tether becoming entangled with environmental objects. 
On the other hand, pre-installation of auxiliary AR-tag (fiducial marker) arrays for robot navigation is not suitable for aging facilities \cite{Duecker2020b,Chen2021}. As mentioned in \cite{legacy_pond},these older facilities often pose extreme health hazards that prohibit human access, making them particularly challenging for the deployment of such markers \cite{Duecker2020b,Chen2021}. 
Among these platforms, the MallARD ASV can operate without a tether and is able to localise itself with millimeter range accuracy.  {However, it is designed solely for surface-level operations.}

Further progress has been made in enhancing the self-localisation capability of underwater robots. For instance, a machine learning enhanced dead reckoning solution was presented in 2019 \cite{Skulstad2019}, however, the meter-level accuracy caused by drift is not suitable for confined spaces. In terms of systems with centimeter-level accuracy, a sensor fusion based framework called SVIN2 was proposed in \cite{Rahman2022} to solve the underwater navigation problem with Simultaneous Localization and Mapping (SLAM). Based on this framework, safe underwater navigation was achieved in \cite{Xanthidis_2020} and \cite{Xanthidis2023a}. However, the experiments in \cite{Xanthidis_2020} and \cite{Xanthidis2023a} demonstrated that the implementation of SVIN2 requires dense and continuous features along the path to update the robot's position, which may not be possible in spent fuel ponds due to visibility constraints, as shown in Fig.~\ref{fig:caves_display1}.

In general, despite the progress and improvement in the hardware design \cite{Griffiths_2016,Duecker2020b,Bhat2020,groves2019m} and software integration \cite{Rahman2022,Xanthidis_2020,Xanthidis2023a} of current underwater robotic platforms, it remains challenging to perform exploration and inspection missions in the target environments using such robots, considering the challenges mentioned in Section \MakeUppercase{\romannumeral 1}.

\subsection{Collective Motion \& Multi-agent Underwater Navigation}

The mechanism of collective motion has been widely studied as a control problem and can be categorised into two types: centralised and decentralised \cite{Hu2021}. In our case, the communication and localization-denied underwater environment necessitate a decentralised approach. A model system for natural and artificial swarms was proposed to describe the collective motion of self-propelled particles in \cite{Ferrante_2013}. More recently, a study analysed the intermittent collective motion of sheep flocks, demonstrating that coordination in these motions results from the propagation of positional information from the leader sheep to all group members through a specific interaction mechanism \cite{GomezNava2022}. These natural collective mechanisms are generally decentralised and have garnered significant attention from researchers in the robotics community. 

Efforts have been made to implement this natural form of connection using physical tethers between robots, aiming to achieve similar collective motion in artificial agents. For instance, a reconfigurable aerial robotic chain was proposed in \cite{Nguyen2020} for heavy-load aerial missions. Compared to aerial vehicles, underwater robots derive greater benefits from tethers. These physical connections between robots have been used for position estimation \cite{Drupt2022} and visual servoing \cite{Laranjeira2020}. Although these systems are inspiring and the connections robust, their maneuverability and safety could be significantly enhanced by virtualising the physical tethers.

To this end, the multi-agent collective motion problem can be simplified to the bearing-only target following problem. Satisfactory solutions have been presented by \cite{He2019,Li2023,Liang2018}, as aerial/ground based bearing-only measurements are relatively more robust compared to underwater systems.
Under the concept of cooperative underwater navigation, communication between robots is generally assumed \cite{Zhang2023} to obtain exteroceptive measurements such as range \cite{Soares_2013} or bearing \cite{Berlinger2021} or both \cite{Kottege2010}. However, these acoustic based solutions will bring serious echo problems \cite{Qu2016} when applied to confined aquatic spaces.  In addition to acoustics, visual servoing based multi-agent underwater navigation has also proven possible, as demonstrated by image-based leader-follower control \cite{Yao2023} and inertial-aided visual servo control \cite{Krupinski_2017}. Although the problem of losing LoS exists \cite{Yao2023}, underwater visual servoing approach provides a potential solution to minimise the reliance on acoustics. 


\subsection{Visual Servoing for Multi-agent Underwater Formation}

Visual servoing methods can be categorised into two types \cite{Lin2021}, 1) Position-Based Visual Servoing (PBVS), which tracks its target by calculating the relative pose based on camera parameters \cite{Chaumette2004}, and 2) Image-Based Visual Servoing (IBVS), which achieves formation by directly controlling the target position within the camera view \cite{Liu2019}. Due to a lower dependency on camera calibration, IBVS is more suitable for underwater environments.

With respect to visual servoing based multi-agent formation, progress has been made in leader-follower control over the past decade. For instance, adaptive pure vision-based leader-follower approaches \cite{Lin2021,Bechlioulis2019a,5299228} have been proposed to achieve the formation of ground vehicles. These approaches require neither communication between the robots, nor the relative position measurement. 
A virtual camera concept was presented in \cite{Lai_2022} to solve the camera tilting problem when controlling an aerial vehicle with visual servoing. Image-based visual servoing for quadrotors has also been applied to track arbitrary flight targets \cite{Wang2023,Li2021c}. With regard to underwater application, studies on visual servoing based multi-agent formation are minimal. An IBVS switchable leader follower control architecture was proposed for heterogeneous multi-agent underwater robotic systems \cite{Yao2023}. However, all these works only consider one-way formation, in which the robot can only be the follower or the leader, despite their roles sometimes being switchable. This could be a cause of unstable formation when the follower is influenced by perturbations, especially in underwater environments. For instance, the loss of LoS was observed in \cite{Yao2023} due to underwater lighting conditions. 

To summarise, former works have only considered the influence of the connection acting on the follower robot. However, to maintain a stable connection under perturbations, which are common in underwater environments, the connection should be mutual. To be specific, the leader robot should also notice the influence of perturbations acting on the follower. As addressed in \cite{GomezNava2022}, alternating the role of leader and follower simultaneously during the formation process is the key to achieving a stable multi-agent navigation. 


\section{PROBLEM FORMULATION}

In this section the generalised underwater navigation problem is formulated as a trajectory tracking problem, followed by assumptions and constraints that address the environmental characteristics. 

\subsection{Problem Statement}
Consider an underwater robot $U$ whose state at time $k$ in the world-fixed coordinate frame $\mathcal{W}$ is given by $\prescript{\mathcal{W}}{}{\bm{x}}_k$, which describes the pose of the robot's body frame $\mathcal{B}$ with respect to the world-fixed frame $\mathcal{W}$ at time $k$, and the state transition is given by

\begin{equation}\label{eq:state_eq}
  \prescript{\mathcal{W}}{}{\bm{x}}_{k} =f(\prescript{\mathcal{W}}{}{\bm{x}}_{k-1}, \bm{u}_{k}, \bm{w}_{k}),
\end{equation}
where $\prescript{\mathcal{W}}{}{\bm{x}} = [x,y,z,\phi, \theta, \psi]^\top$, and $x,y,z$ denote the robot's position while the angles $\phi, \theta, \psi$ represent its orientation around $x$, $y$, and $z$, respectively. Control input is represented by $\bm{u}_k$, whilst $\bm{w}_k$ denotes disturbances acting on the robot. 

The state trajectory is given by $\mathcal{X}=(\prescript{\mathcal{W}}{}{\bm{x}}_{1},\prescript{\mathcal{W}}{}{\bm{x}}_{2},\dots,\prescript{\mathcal{W}}{}{\bm{x}}_{n})$ and the observation of the state at time $k$ is

\begin{equation}\label{eq: measure_eq}
  \prescript{\mathcal{W}}{}{\bm{z}}_{k} =g(\prescript{\mathcal{W}}{}{\bm{x}}_{k-1}, \bm{v}_{k}),
\end{equation}
where $\prescript{\mathcal{W}}{}{\bm{z}}$ is the estimate of the state obtained by self-localisation and $\bm{v}$ is the measurement noise. 

Let $\mathcal{P}=(\bm{p}_1, \bm{p}_2, \dots, \bm{p}_n)$ be the target trajectory in the navigation mission, consisting of a set of target setpoints. The Euclidean distance between the target trajectory $\mathcal{P}$ and the actual trajectory $\mathcal{X}$ of the robot at time $k$ is given by 

\begin{equation}\label{eq: overlap_eq}
  \prescript{\mathcal{W}}{}{D}^{\mathcal{P}_k}_{\mathcal{X}_k} = ||\mathcal{P}_k-\mathcal{X}_k(\bm{u}_{k-1})||.
\end{equation}

{Therefore, the mission objective is formulated as designing the controller that aims to reduce $\prescript{\mathcal{W}}{}{D}^{\mathcal{P}_k}_{\mathcal{X}_k}$ at each time step:
\begin{equation}\label{eq: target}
  \bm{u}_k = \mathcal{G}(\prescript{\mathcal{W}}{}{D}^{\mathcal{P}_k}_{\mathcal{X}_k}),
\end{equation}
the controller adjusts $\bm{u}_k$ in a way that progressively reduces this distance, thereby aligning the actual trajectory of the robot with the desired trajectory.}



\subsection{Assumptions \& Constraints}
To solve the autonomous underwater navigation problem defined by (\ref{eq: target}), two assumptions are imposed:

\textit{Assumption 1:} {Under sufficiently tightly bounded $\bm {w}_{i+1}$,} the state transition between any two consecutive setpoints 
$\bm{p}_i\rightarrow\bm{p}_{i+1}$ on the trajectory $\mathcal{P}$ is within the maneuvering capability of the robot. Therefore, there is always a control input $\bm{u}_{i+1}$ for robot $U$ such that the following equation holds,

\begin{equation}\label{eq:assumption1}
  \prescript{\mathcal{W}}{}{\bm{p}}_{i+1} =f(\prescript{\mathcal{W}}{}{\bm{p}}_{i}, \bm{u}_{i+1},\bm {w}_{i+1}).
\end{equation}

\textit{Assumption 2:} The velocity of the underwater robot is bounded, i.e. there exists a real number $M>0$ such that the following inequalities hold for all $k$,

\begin{equation}\label{eq:assumption2}
|| {\prescript{\mathcal{W}}{}{\bm{\dot x}}_{k}}|| \leq M,
\end{equation}
where  $||{\prescript{\mathcal{W}}{}{\bm{\dot x}}}||$ is the magnitude of the velocity. 

Considering the environmental characteristics, the following constraints are introduced:
\subsubsection{Limited Communication}
For underwater robot $U$, both robot-to-robot and robot-to-ground station communication are denied. 

\subsubsection{Partial-State Localisation}
Measurable self-state, obtained by means of self-localisation, is limited to depth-$z$, roll angle-$(\phi)$, and pitch angle-$(\theta)$: 

\begin{equation}\label{eq:constraints2}
\prescript{\mathcal{W}}{}{\bm{z}} = [z,\phi, \theta]^\top.
\end{equation}

\section{Virtual Elastic Tether}
Generally, addressing the trajectory tracking problem (\ref{eq: target}) requires either accurate and full-state self-localisation in order to provide control feedback,
which conflicts with the localisation constraint, or the need for remote intervention to correct the robot's trajectory during the mission, which violates the communication constraint.

Tether-guided motion control is prevalent across various domains, ranging from commonplace activities such as walking a dog or flying a kite, to more specialized scenarios such as human chain rescue operations in coastal areas. The unifying aspect of these examples lies in the interaction between two agents: one agent lacks the ability to accomplish the desired motion independently, while the other provides guidance via a robust connection between them. 

To this end, we consider a solution that uses a robot  $S$ operating on the surface of the water as a navigation aid to `guide' the underwater robot $U$ to follow its desired trajectory. 

In order to account for the scenario where the two aforementioned robots are considered,  the model of the system (\ref{eq:state_eq}) can be extended to a multi-agent case 

\begin{subequations}\label{eq:system_model}
\begin{align}
   U: \qquad & \prescript{\mathcal{W}}{}{\bm{x}}^U_{k} = f^U(\prescript{\mathcal{W}}{}{\bm{x}}^U_{k-1}, \bm{u}^U_{k}, \bm{w}^U_{k}) \\[5pt]
   S : \qquad & \prescript{\mathcal{W}}{}{\bm{x}}^S_{k} = f^S(\prescript{\mathcal{W}}{}{\bm{x}}^S_{k-1}, \bm{u}^S_{k}, \bm{w}^S_{k}),
\end{align}
\end{subequations}
where the superscripts $U$ and $S$ indicate the underwater and surface robots, therefore $\prescript{\mathcal{W}}{}{\bm{x}}^U = \left[x^U,y^U,z^U,\phi^U, \theta^U, \psi^U \right]^\top$ and $\prescript{\mathcal{W}}{}{\bm{x}}^S = \left[x^S,y^S,\psi^S \right]^\top$ represent the states of the robots $U$ and $S$ in the world-fixed frame $\mathcal{W}$, respectively.

Since self-localisation problem in air has been relatively well solved compared to its underwater counterpart \cite{groves2021}, it is reasonable to assume that $\prescript{\mathcal{W}}{}{\bm{z}}^S  \approx \left[x^S,y^S,\psi^S \right]^\top$. Therefore, the underwater navigation problem can be divided into two parts:
\begin{enumerate}
    \item guiding robot $U$ to the desired depth ($Z_d^{U}$), whilst maintaining certain roll ($\phi$) and pitch ($\theta$) angles;
    \item directing robot $U$ to follow the projection of robot $S$'s trajectory at the depth ($Z_d^{U}$). 
\end{enumerate}
Note that the second problem can be reformulated as a leader-follower problem. However, unlike traditional leader-follower approaches, the method proposed in this paper explicitly accounts for the connectivity between the leader and the follower. Such connection between the two robots is modeled through VET, which aims to achieve leader-follower formation by having both agents actively contribute to ensuring their virtual connection is maintained at all times. 

To model the virtual tether connection, we first extend the observation of the system state (\ref{eq: measure_eq}) to 
  {\begin{subequations}\label{eq:system_measure}
\begin{align}
 U: \qquad &   \prescript{\mathcal{W}}{}{\bm{z}}^U_{k} =g^U(\prescript{\mathcal{W}}{}{z}^U_{k-1}, \prescript{\mathcal{W}}{}{\phi}^U_{k-1}, \prescript{\mathcal{W}}{}{\theta}^U_{k-1}, \xi ^{U \rightarrow S}_{k-1},  \bm{v}^U_{k}) \label{eq:system_measure_a} \\[5pt]
   S : \qquad & \prescript{\mathcal{W}}{}{\bm{z}}^S_{k} =g^S(\prescript{\mathcal{W}}{}{x}^S_{k-1}, \prescript{\mathcal{W}}{}{y}^S_{k-1}, \prescript{\mathcal{W}}{}{\psi}^S_{k-1}, \xi ^{S \rightarrow U}_{k-1}, \bm{v}^S_{k}), \label{eq:system_measure_b}
\end{align}
\end{subequations}
where} $\prescript{\mathcal{W}}{}{\bm{z}}^U$ and $\prescript{\mathcal{W}}{}{\bm{z}}^S$ represent the measured states of robots \(U\) and \(S\) respectively, while \(g^U\) and \(g^S\) are their associated measurement functions. Moreover, $\xi ^{U \rightarrow S}$ and $\xi ^{S \rightarrow U}$ represent the measurements of relative tether states, denoted as $\xi = \left[ \xi^{U \rightarrow S}\xi^{S \rightarrow U}  \right]^\top$. Introduced by VET, the tether state serves as an additional state for each robot, indicating proximity from one robot to the other. This state $\xi$ is a function of the control input $\bm{u}$ as well as the disturbance $w$, i.e. $\xi(\bm {u},\bm w)$.

It should be noted that the exact representation of $\xi(\bm u)$ can vary depending on the application context. For instance, in scenarios where distance is measurable, $\xi$ might denote the projected distance between two robots.  Conversely, when visual markers are utilised, the dimension or positioning of these markers within the image plane might be adopted to represent $\xi$. Therefore, for the sake of generality, the explicit form of $\xi(\bm u)$ is not provided in this section (implementation example shown in Fig.~\ref{fig:adaptive_elastic_force}).


Consider the trajectories of the two robots $\mathcal{X}^U$ and $\mathcal{X}^S$, the mission objective, as given in (\ref{eq: target}), can be re-formulated as a leader-follower problem that aims to minimise Euclidean distance between $\mathcal{X}^U$ and $\mathcal{X}^S$ by controlling the relative tether states at each time step:
{\begin{subequations}\label{eq:system_target}
\begin{align}
\prescript{}{}{\bm{u}}_k = \mathcal{G}(\prescript{\mathcal{W}}{}{D}^{\mathcal{X}^U_{z=0}}_{\mathcal{X}^S}) & = \mathcal{F}(\prescript{e}{}{\bm {u}_k},\prescript{\xi}{}{\bm {u}_k}) \label{eq:system_target_1} \\[10pt]
\text{where:} \quad &\prescript{\xi}{}{\bm{u}_k} \propto ||\xi_k - \xi_{d}||, \label{eq:system_target_3} \\[10pt]
\textbf{s.t. }  \forall k , & \; \xi_{min} < \xi_{k} < \xi_{max}, \label{eq:system_target_2}    
\end{align}
\end{subequations}
where} $\prescript{\mathcal{W}}{}{D}^{\mathcal{X}^U_{z=0}}_{\mathcal{X}^S}$ denotes the Euclidean distance between the projected trajectory of robot $U$ and robot $S$ on the plane $z=0$ at time $k$, while $\mathcal{F}$ denote the overall time-varying multi-objective controllers for the robots. Meanwhile, $\prescript{e}{}{\bm{u}}_k = \left[ \prescript{e}{}{\bm{u}}^U_k, \prescript{e}{}{\bm{u}}^S_k \right]^\top$ represent the control inputs for sub-tasks (such as depth control or tilt angles control) of robot $U$ and $S$, and $\prescript{\xi}{}{\bm{u}}_k = \left[ \prescript{\xi}{}{\bm{u}}^U_k, \prescript{\xi}{}{\bm{u}}^S_k \right]^\top$ represents the control inputs that drive the relative tether state $\xi_k=\left[\xi ^{U \rightarrow S}_k,\xi ^{S \rightarrow U}_k \right]^\top$ towards its desired state $\xi_{d}=\left[\xi ^{U \rightarrow S}_{d},\xi ^{S \rightarrow U}_{d} \right]^\top$, which is user-specified value that can be set accordingly depending on a particular application. The constraint (\ref{eq:system_target}c) accounts for the \textbf{elasticity} of the virtual tether.


 Therefore, the solution to the navigation problem in (\ref{eq:system_target_1}) can be expressed as the feedback control law given by

\begin{subequations}\label{eq:feedback}
\begin{align}
   U: \qquad  &   \prescript{\mathcal{\xi}}{}{\bm{u}}^U_{k} =\Xi^{U}(\xi^{U \rightarrow S}_{k-1}, \bm{ w}_k^U) \\[5pt]
   S : \qquad &  \prescript{\mathcal{\xi}}{}{\bm{u}}^S_{k} =\Xi^{S}(\xi^{S \rightarrow U}_{k-1}, \bm{w}_k^S),
\end{align}
\end{subequations}
where $\Xi^{U}$ and $\Xi^{S}$ denote the VET-based control law for robots $U$ and $S$, respectively. Additionally, to ensure the robots are acting simultaneously to maintain the connection and ensure convergence to $\xi = \xi_{d}$, the following two conditions are imposed

\begin{subequations}\label{eq:connectivity}
\begin{align}
   \textbf{s.t. } &\frac{\Xi^U(\xi^U-\xi^U_\text{des})}{||\Xi^U(\xi^U-\xi^U_\text{des})||} \approx- \frac{\Xi^S(\xi^S-\xi^S_\text{des})}{||\Xi^S(\xi^S-\xi^S_\text{des})||}\label{eq:connectivity_1}\\[5pt]
   &\Xi^U(\xi^{U \rightarrow S}_\text{des}) \approx \Xi^S(\xi^{S \rightarrow U}_\text{des}) \approx 0 \label{eq:connectivity_2}
\end{align}
\end{subequations}
where (\ref{eq:connectivity_1}) ensures the control inputs for the two robots are symmetrical about $\xi = \xi_{d}$. {This symmetry means that the direction and magnitude of the control efforts of the two robots are balanced in a way that they counteract each other's actions (approximate equal and opposite reactions). Such a balance is crucial as the objective is to maintain a certain relational state ($\xi$) and connection between the robots.} Additionally, (\ref{eq:connectivity_2})  guarantees that the control inputs converge to about 0 when the desired tether state is reached. Together (\ref{eq:connectivity_1}) and (\ref{eq:connectivity_2}) describe the \textbf{connectivity} of the virtual tether. 



In summary, considering the navigation problem for the multi-agent system mentioned above, each robot in the system has partial-state measurements, 

\begin{align}\label{eq:system_sum}
dim(\bm{z}_k^U) < dim(\bm{x}_k^U) = dim(\bm{z}_k^U) + dim(\bm{z}_k^S)
\end{align}
where $dim($·$  )$ denotes the dimensionality of the state vectors. The observation vector $\bm{z}^U = [z, \phi,\theta]^\top$ is supplemented with unobservable states $[x, y, \psi]^\top$ using $\bm{z}_k^S$ of robot $S$. Incorporating with (\ref{eq:feedback}), the controller for the multi-agent system is given by

\begin{subequations}\label{eq:system_controller}
\begin{align}
  U:&   \bm{u}_{k}^{U}=\mathcal{F}^U(E^U(\bm{z}^U_{k-1}),\Xi^{U}(\xi^{U \rightarrow S}_{k-1})) \\[5pt]
  S:& \bm{u}_{k}^{S}=\mathcal{F}^S(E^S(\bm{z}^S_{k-1}),\Xi^{S}(\xi^{S \rightarrow U}_{k-1})),\\[5pt]
  \textbf{s.t. }& \ref{eq:system_target_2},\ref{eq:connectivity_1},\ref{eq:connectivity_2} \label{eq:system_controller_3}
\end{align}
\end{subequations}
where $E^U$, $E^S$ represent the controllers for sub-tasks of robot $U$ and $S$. In this underwater-surface multi-agent scenario, $E^U$ denotes the depth, roll, and pitch controller for the underwater robot $U$, while $E^S$ represents the 2-D path follower for the surface robot $S$. Properties of \textbf{elasticity} and \textbf{connectivity} of VET are ensured by imposing (\ref{eq:system_controller_3}).

Now, the remaining task is to design suitable functions $\mathcal{F}$, $E$, and $\Xi$ within the framework of (\ref{eq:system_controller}a-c). This feedback-based multi-agent formation framework can be adapted to various scenarios. For example, incorporating a local planner for obstacle avoidance into function $E$ transforms the system into a cooperative obstacle avoidance system with robust leader-follower formation capabilities. It is important to note that the measurements of $\xi$ are not restricted to any specific sensor type. They can be acoustic-based bearings, odometry-based distances, or visual-based relative positions, among others. The following section will provide an example of a visual-based VET implementation in an extreme aquatic environment where conventional systems fail to perform adequately.


\section{Visual-based VET implementation of AUV-ASV Leader-follower Formation}

\begin{figure}[tpb]
      \centering
       \includegraphics[width=0.43\textwidth]{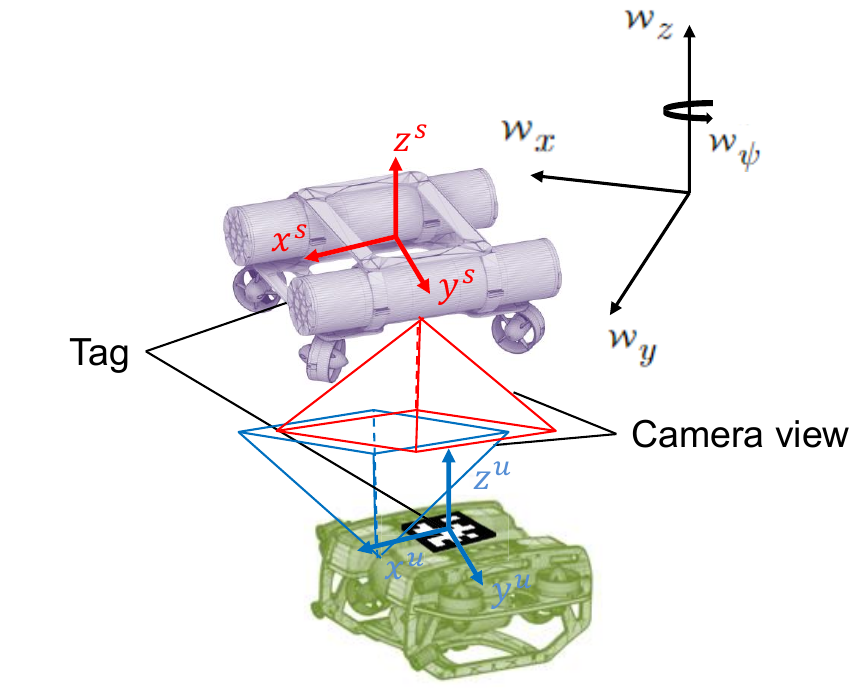}
      \caption{Coordinate system of the proposed multi-agent platform.
      }
      \label{fig:caves_model}
  \end{figure}

In order to accomplish the underwater exploration mission in confined water ponds, where underwater communication and localisation are severely limited, this paper proposes a cooperative aquatic vehicle exploration system, abbreviated as CAVES, that incorporates both an AUV and an ASV. The system is comprised of two robots: the BlueROV2 AUV and the MallARD ASV \cite{Groves2019}. While the AUV lacks self-localisation capabilities, the ASV can only localise itself on the water surface. As per (\ref{eq:system_measure}a,b), the BlueROV2 is equipped with a depth sensor and an IMU (Inertial Measurement Unit), while the MallARD utilises a 2-D LiDAR for surface localisation. The AUV is guided by the ASV, and their {formation} is stabilized by utilising VET to ensure robust guidance. Both robots are fitted with cameras, facing each other, to provide the measurement of the virtual tether state, $\xi$, as illustrated in Fig.~\ref{fig:caves_model}.


\subsection{System Model}

Both the BlueROV2 and MallARD are omnidirectional in terms of maneuverability, therefore the kinematic model for each robot can be given as
\begin{subequations}\label{eq:sys_kinematic}
\begin{align}
   U:  & \prescript{\mathcal{W}}{}{\bm{\dot x}}^U = \bm{J^U} \bm{\nu}^U \label{eq:sys_kinematic1}\\[5pt]
   S:  & \prescript{\mathcal{W}}{}{\bm{\dot x}}^S = \bm{J^S} \bm{\nu}^S,\label{eq:sys_kinematic2}
\end{align}
\end{subequations}
where $\bm{J}^U\in \mathbb{R}^{6 \times 6}$ and $\bm{J}^S\in \mathbb{R}^{3 \times 3}$ denote the velocity Jacobians (\cite{groves2019m,Yao2023}) for AUV and ASV from the robot frame $\mathcal{B}$ to the world-fixed frame $\mathcal{W}$. The body frame velocities for AUV and ASV are represented by $\bm{\nu}^U\in \mathbb{R}^{6}$ and $\bm{\nu}^S\in \mathbb{R}^{3}$, respectively. 

Following Fossen \cite{Fossen1995}, the dynamic model for both robots is approximated by 
{\begin{equation}\label{eq:sys_dynamic}
\begin{array}{rcl}
  \bm M\bm{\dot \nu} + \bm H(\bm\nu)\bm\nu &=& \bm\tau, \\
  \text{where} \quad \bm H(\bm\nu)&=&\bm C(\bm\nu)+ \bm D(\bm\nu)
  \end{array}
\end{equation}
where} {$\bm\nu=\bm\nu^U$ for underwater robot, and $\bm\nu=\bm\nu^S$ for surface robot.} The matrices $\bm M$, $\bm C$, and $\bm D$ represent the individual inertia, Coriolis rigid body effects, and damping for each robot, respectively. These matrices can be determined using the approach described in \cite{groves2019m}. The vector $\bm{\tau}=[\bm \tau^U, \bm \tau^S]^\top$ denotes the thrust forces $\bm {f}_i$ and moments $\bm {m}_i$ generated by the individual thruster $i$, which can be converted to the resultant effects from control inputs:
\begin{equation}\label{eq:sys_inputs}
\begin{array}{rcl}
\bm\tau&= &\sum\limits_{i=1}^{n} \left[ \begin{array}{c}
                            \bm {f}_i\\
                           \bm {m}_i + \bm {r}_i \times \bm {f}_i
                            \end{array} \right]\\[10pt]
&=&\Lambda_{K|6 \times 6} [\underbrace{u_{k|x}, u_{k|y}, u_{k|z}, u_{k|\phi}, u_{k|\theta}, u_{k|\psi}}_{u_k}]^\top             
\end{array},
\end{equation}
where $n$ is determined by the thruster layout of the robots: $n=8$ for BlueROV2 in its heavy configuration and $n=4$ for MallARD. The detailed forms of diagonal linear force coefficient matrices $\Lambda^U_{K}$ and $\Lambda^S_{K}$ allow convenient gain tuning of the individual components. Note that for MallARD, $\bm{u}^S_k$ yields to $[{u^S_{k|x}, u^S_{k|y}, u^S_{k|\psi}}]^\top$, corresponding to its kinematics (\ref{eq:sys_kinematic2}).


\subsection{Visual VET for Underwater Leader-Follower Formation}
The sub-task function $E$ for each of the robots can be expressed as the following feedback control law:
\begin{equation}\label{eq:E_design}
\begin{array}{rc}
   E^U: &  \left[0, 0, u^U_{k|z}, u^U_{k|\phi},u^U_{k|\theta}, 0 \right]^\top = \bm{K}^U_p \bm{e}^U_E + \bm{K}^U_d {\bm{\dot e}^U_E} \\[5pt]
    E^S: &  \left[  {u^S_{k|x}, u^S_{k|y}, u^S_{k|\psi}}\right]^\top = \bm{K}^S_p \bm{e}^S_E + \bm{K}^S_d {\bm{\dot e}^S_E}
\end{array},
\end{equation}
where $\bm{e}^U_E$ and $\bm{e}^S_E$ denote the error between desired and measured states, respectively. The matrices $\bm{K}_p$ and $\bm{K}_d$ are diagonal and represent controller gains corresponding to individual elements of the error vector or its time derivative. As per (\ref{eq:constraints2}), in terms of the underwater robot, only the states of depth-$z$, roll angle-$\phi$, and pitch angle-$\theta$ are measurable. {This implies that the corresponding components for the non-measurable states $x^U$, $y^U$, and $\psi^U$  in the matrices $\bm{K}_p^U$ and $\bm{K}_d^U$ are set to zero, as these states do not contribute to the control input. Therefore, we have:} 
{\begin{equation}\label{eq:error=0}
\left\{
\begin{array}{cc}
    (\bm{K}_p^U)_{1,1} = (\bm{K}_p^U)_{2,2} = (\bm{K}_p^U)_{6,6} = 0 \\[5pt]
    (\bm{K}_d^U)_{1,1} = (\bm{K}_d^U)_{2,2} = (\bm{K}_d^U)_{6,6} = 0
\end{array}.
    \right.
\end{equation}}

Following (\ref{eq:system_model}) and (\ref{eq:error=0}), we define the error vector as 

\begin{equation}\label{eq:E_e}
\begin{array}{rrl}
   U: &  \bm{e}^U_E &=  \left[ \begin{array}{c}
                                        0\\
                                        0\\
                              z^U_d - z^U_k\\
                           \phi^U_d - \phi^U_k\\
                           \theta^U_d- \theta^U_k\\
                                        0
                            \end{array} \right]  \\[16pt]
   S: &   \bm{e}^S_E &=  \left[ \begin{array}{c}
                              x^S_d - x^S_k\\
                           y^S_d - y^S_k\\
                           \psi^S_d - \psi^S_k
                            \end{array} \right] 
\end{array},
\end{equation}
where {$z^U_d$, $\phi^U_d$ and $\theta^U_d$ are desired states of the underwater robot, $z^U_k$, $\phi^U_k$ and $\theta^U_k$} are given by the depth sensor and IMU mounted on BlueROV2, following the localisation assumption (\ref{eq:constraints2}). {Note that in practice, to avoid the unwinding problem—where the control system might unnecessarily rotate through a larger angle due to the periodic nature of angles—all angular measurements and commands are normalized to be within the range of  $-\pi$ and $\pi$.} For MallARD, the {$x^S_k$, $y^S_k$ and $\psi^S_k$ }are given by an onboard 2-D LiDAR, which feeds into $E^S$, formulating the surface trajectory tracking controller \cite{Groves2019}.

  \begin{figure}[tpb]
      \centering
       \includegraphics[width=0.48\textwidth]{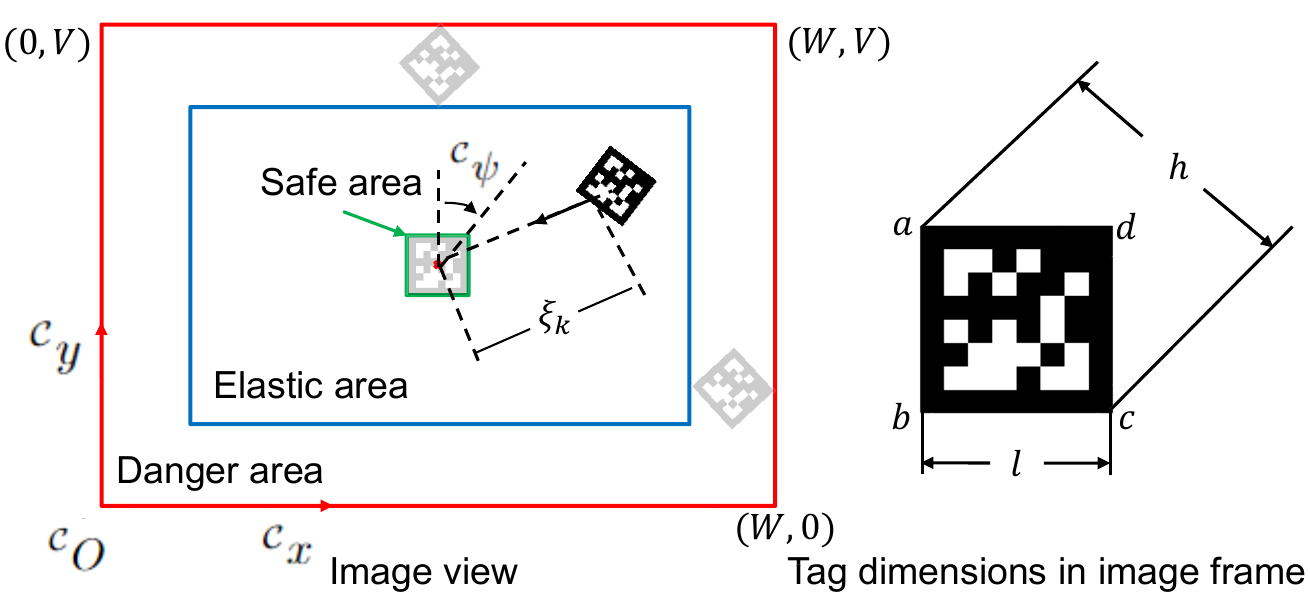}
      \caption{At time $k$, the left image shows three different effective areas in the image view, with the size of these areas adapting to the measurement of the actual tag size in the image frame, $\xi_k$ represents the tether state; the right image shows the tag dimensions in the image frame. Note that $l$ and $h$ are calculated based on the positions of the corners ($a$, $b$, $c$, and $d$) in the image frame. $V$ and $W$ are determined by the camera's pixels.  
      }
      \label{fig:adaptive_elastic_force}
  \end{figure}

As discussed in Section \MakeUppercase{\romannumeral 1}, to reduce the reliance on accurate pose estimation and increase the robustness to inaccuracies in image measurements, IBVS is selected as VET measurement basis. While various object detection methods exist, ARTag-based approach \cite{Wang2016} is adopted in this paper to obtain $\xi_k$. To design the VET control function $\Xi$, the tag dimension obtained in the image plane was used to divide the camera view into three areas based on the proximity to losing LoS (as shown in Fig.~\ref{fig:adaptive_elastic_force}),
{\begin{equation}\label{eq:elastic_image}
\left\{
\begin{array}{rl}
   \Omega _\text{safe} \stackrel{\Delta}{=} & \left\{(x,y)\in \mathbb{R}^2:\right. \\ [5pt]
   &\left. x \in (\frac{W- l}{2}, \frac{W+ l}{2}) \land y \in (\frac{V- l}{2}, \frac{V+l}{2})\right\} \\[5pt]
   \Omega _\text{elastic} \stackrel{\Delta}{=}& \left\{(x,y)\in \mathbb{R}^2: \right. \\ [5pt]
  &\left. x \in (h, V- h) \land y \in (h, W- h) \right\} \cap \bar{\Omega} _\text{safe}\\[5pt]
   \Omega _\text{danger} \stackrel{\Delta}{=}& \left\{(x,y)\in \mathbb{R}^2: \right. \\ [5pt]
  &\left. x \in (0, W) \land y \in (0, V)\right\} \cap \bar{\Omega} _\text{elastic} \cap \bar{\Omega} _\text{safe}.
   \end{array}
\right.
\end{equation}}

The set definition (\ref{eq:elastic_image}) ensures the elasticity \textbf{elasticity} of VET, where $V$ and $W$ are determined by the camera resolution, $l$ and $h$ denote tag dimensions in the camera frame, $\prescript{\mathcal{C}}{}{\psi}$, given by Apriltag library \cite{Wang2016}, denotes the yaw angle error in the camera frame. Assuming that the desired roll angle-$(\phi^U_d)$ and pitch angle-$(\theta^U_d)$ are 0 and with the centerline of the tag aligned with the mounted robot's centerline, then $\prescript{\mathcal{C}}{}{\psi}_k \approx \prescript{\mathcal{W}}{}{\psi}^U_k-\prescript{\mathcal{W}}{}{\psi}^U_k$.
  
It is important to note that with the same robot and tag sizes, the values of $l$ and $h$ will vary based on the camera's tilt angle and the actual distance between the robots. Therefore, we use $\bar l$ and $\bar h$ as adaptive substitutes for $l$ and $h$. These values are calculated based on the measured positions of the four corners ($a$, $b$, $c$, and $d$) in the image frame $\mathcal{C}$, as shown in Fig.~\ref{fig:adaptive_elastic_force},
{\begin{equation}\label{eq:dim_image}
\begin{array}{c}
(\prescript{\mathcal{C}}{}{\bar x}, \prescript{\mathcal{C}}{}{\bar y}) = \frac{1}{4}(\prescript{\mathcal{C}}{}{x}_a+\prescript{\mathcal{C}}{}{x}_b+\prescript{\mathcal{C}}{}{x}_c+\prescript{\mathcal{C}}{}{x}_d, \prescript{\mathcal{C}}{}{y}_a+\prescript{\mathcal{C}}{}{y}_b+\prescript{\mathcal{C}}{}{y}_c+\prescript{\mathcal{C}}{}{y}_d)\\[5pt]
\left\{
\begin{array}{rl}
\bar l &= \frac{1}{4} (ab+bc+cd+ad)\\[5pt]
\bar h &= \frac{1}{2} (ac+bd), 
\end{array}
\right.
\end{array}
\end{equation}
where $(\prescript{\mathcal{C}}{}{\bar x}, \prescript{\mathcal{C}}{}{\bar y})$ represents the centre of the tag ($C_\text{tag}$) in the camera view, and $ab$, $bc$, $cd$, $ad$, $ac$, and $bd$ denote the length of segments connecting two corners.} That is to say, when the robot's centre is within the safe area, this indicates that the VET is in a relaxed state with a low likelihood of losing LoS. In this scenario, there is no elastic force, and the robot should prioritise other tasks. Conversely, if the robot is in the danger area, the VET is in a taut state, with the virtual force at its maximum, and the robot should prioritise the leader-follower mission. The elastic area is the primary functional region where the elastic force is generated based on the robot's position and velocity within the image frame. {Let the centre of the image view $\bm c_\text{image}(\frac{W}{2}, \frac{V}{2})$,}  to calculate control input in the camera frame $\prescript{\mathcal{C}}{}{\bm{u}}_k=[\prescript{\mathcal{C}}{}{u}_{k|x}, \prescript{\mathcal{C}}{}{u}_{k|y}, \prescript{\mathcal{C}}{}{u}_{k|\psi}]^\top$, we design the function $\Xi$ based on LoS awareness mechanism as in \textbf{Algorithm \MakeUppercase{\romannumeral 1}}.

  \begin{table}[htbp]
  \centering
  \label{tab:algorithm}
  \begin{tabular}{rp{6.7cm}}   
  \toprule    
  {\textbf{Algorithm \MakeUppercase{\romannumeral 1}:}}&{{\textbf{LoS Awareness Mechanism}} }                                              \\[3pt]
    \toprule  
      \textbf{Input:} &Position of the tag mounted on the robot in image frame $\prescript{\mathcal{C}}{}{\psi}$, $(\prescript{\mathcal{C}}{}{x}_a, \prescript{\mathcal{C}}{}{y}_a)$, $(\prescript{\mathcal{C}}{}{x}_b, \prescript{\mathcal{C}}{}{y}_b)$, $(\prescript{\mathcal{C}}{}{x}_c, \prescript{\mathcal{C}}{}{y}_c)$ and $(\prescript{\mathcal{C}}{}{x}_d, \prescript{\mathcal{C}}{}{y}_d)$    \\[3pt]
        \textbf{1:}&  \textbf{If} $\text{Detection}_\text{tag} = \text{True}$ \\[3pt]
        \textbf{2:}&   \quad  \textbf{Calculate:} $ \bm c_\text{tag} \gets (\prescript{\mathcal{C}}{}{\bar x}, \prescript{\mathcal{C}}{}{\bar y})$, $ {\bm{\dot c}}_\text{tag} \gets (\frac{d \prescript{\mathcal{C}}{}{\bar x}}{dt} , \frac{d \prescript{\mathcal{C}}{}{\bar y}}{dt} ) $       \\[3pt]
        \textbf{3:}&  \quad \textbf{If} $\bm c_\text{tag} \in \Omega _\text{safe} $\textbf{:} \\[3pt]
        \textbf{4:}&  \qquad $[\prescript{\mathcal{C}}{}{u}_{k|x}, \prescript{\mathcal{C}}{}{u}_{k|y}]^\top = K_{s|p}[\bm{c_\text{tag}}-\bm c_\text{image}]^\top$ \\[3pt]
        \textbf{5:} &  \quad \textbf{Else if} $\bm c_\text{tag} \in \Omega _\text{elastic}$ \textbf{:}\\[3pt]
        \textbf{6:} &    \qquad  $[\prescript{\mathcal{C}}{}{u}_{k|x}, \prescript{\mathcal{C}}{}{u}_{k|y}]^\top = K_{e|p}[\bm{c_\text{tag}}-\bm c_\text{image}]^\top + K_{e|d} \bm{\dot {c}_\text{tag}} $   \\[3pt]
        \textbf{7:} &   \quad  \textbf{Else: } \\[3pt]
        \textbf{8:}  &   \qquad $[\prescript{\mathcal{C}}{}{u}_{k|x}, \prescript{\mathcal{C}}{}{u}_{k|y}]^\top = [\prescript{\mathcal{C}}{}{u}_{\text{max}|x}, \prescript{\mathcal{C}}{}{u}_{\text{max}|y}]^\top$   \\ [3pt]     
        \textbf{9:}&     \quad \textbf{End if }    \\ [3pt]
        \textbf{10:}&     \textbf{End if }    \\[3pt]
        \textbf{11:}&    \quad  $\prescript{\mathcal{C}}{}{u}_{k|\psi} = K_{\psi}\prescript{\mathcal{C}}{}{\psi}_k $    \\ [3pt]
      \textbf{Output:} & Control input  $\prescript{\mathcal{C}}{}{u}_k$  \\
             \bottomrule 
  \end{tabular} 
  \end{table}
In function $\Xi$, the proportional gains meet $K_{s|p} < K_{e|p}$, the max input $ [\prescript{\mathcal{C}}{}{u}_{max|x}, \prescript{\mathcal{C}}{}{u}_{max|y}]^\top$ is determined by the hardware constraints of the robot, it meets $[\prescript{\mathcal{C}}{}{u}^U_{max|x}, \prescript{\mathcal{C}}{}{u}^U_{max|y}]^\top = [\prescript{\mathcal{C}}{}{u}^S_{max|x}, \prescript{\mathcal{C}}{}{u}^S_{max|y}]^\top$, corresponding to assumption 1 (\ref{eq:assumption1}). {Therefore, when the desired tether state $\xi_\text{des}$ is achieved, such that $\bm{c_\text{tag}}=\bm c_\text{image}$, and recalling the feedback loop defined in (\ref{eq:feedback}), we can then formulate the function $\Xi^U$ and $\Xi^S$ as follows:} 
\begin{equation}\label{eq:Xi}
\begin{array}{rc}
   \Xi^U:& \left[ \prescript{\xi}{}{u^U_{k|x}} , \prescript{\xi}{}{u^U_{k|y}}, 0, 0, 0, \prescript{\xi}{}{u^U_{k|\psi}}\right]^\top = \prescript{\mathcal{C}}{\mathcal{B}}{\bm{T}}^U  \prescript{\mathcal{C}}{}{\bm{u}}^U_k\\ [5pt]
  \Xi^S:& \left[ \prescript{\xi}{}{u^S_{k|x}} , \prescript{\xi}{}{u^S_{k|y}} ,\prescript{\xi}{}{u^S_{k|\psi}} \right]^\top = \prescript{\mathcal{C}}{B}{\bm{T}}^S \prescript{\mathcal{C}}{}{\bm{u}}^S_k ,

\end{array}
\end{equation}
where $\prescript{\mathcal{C}}{\mathcal{B}}{\bm{T}}^U, \prescript{\mathcal{C}}{\mathcal{B}}{\bm{T}}^S \in SO(3)$ denote the transformation matrices from the camera frame $\mathcal{C}$ to robot body-fixed frame $\mathcal{B}$ of BlueROV and MallARD, respectively. These transformation matrices are determined by the physical mounting position of the camera robots. Let $\prescript{\mathcal{B}}{}{\bf{c}}^U_C = \left[\prescript{\mathcal{B}}{}{x}^U_C,\prescript{\mathcal{B}}{}{y}^U_C,\prescript{\mathcal{B}}{}{z}^U_C \right]^\top$, $\prescript{\mathcal{B}}{}{\bf{c}}^U_T =\left[\prescript{\mathcal{B}}{}{x}^U_T,\prescript{\mathcal{B}}{}{y}^U_T,\prescript{\mathcal{B}}{}{z}^U_T \right]^\top$ and $\prescript{\mathcal{B}}{}{\bf{c}}^S_C = \left[\prescript{\mathcal{B}}{}{x}^S_C,\prescript{\mathcal{B}}{}{y}^S_C,\prescript{\mathcal{B}}{}{z}^S_C \right]^\top$, $\prescript{\mathcal{B}}{}{\bf{c}}^S_T  = \left[\prescript{\mathcal{B}}{}{x}^S_T,\prescript{\mathcal{B}}{}{y}^S_T,\prescript{\mathcal{B}}{}{z}^S_T \right]^\top$ be the centre of camera and tag that mounted on BlueROV and MallARD. To ensure that VET simultaneously directs both robots towards the same direction and deactivates at the desired state, the following equation must hold:
{\begin{equation}\label{eq:image_connectivity}
\begin{array}{rl}
    \prescript{\mathcal{B}}{\mathcal{W}}{\bm{T}}^U  \left[ \begin{array}{c}
                                        \prescript{\mathcal{B}}{}{\bf{c}}^U_C - \prescript{\mathcal{B}}{}{\bf{c}}^U_T\\
                                        1
                            \end{array} \right] =  - \prescript{\mathcal{B}}{\mathcal{W}}{\bm{T}}^S \left[ \begin{array}{c}
                                        \prescript{\mathcal{B}}{}{\bf{c}}^S_C - \prescript{\mathcal{B}}{}{\bf{c}}^S_T\\
                                        1
                            \end{array} \right]
\end{array},
\end{equation}
where $\prescript{\mathcal{B}}{\mathcal{W}}{\bm{T}}^U, \prescript{\mathcal{B}}{\mathcal{W}}{\bm{T}}^S \in SE(3)$} represent the transformation matrices from the robot body-fixed frame $\mathcal{B}$ to the world-fixed frame $\mathcal{W}$ of BlueROV and MallARD, respectively. The above equation denotes the \textbf{connectivity} of VET. 

By combining (\ref{eq:E_design}), (\ref{eq:Xi}) and (\ref{eq:image_connectivity}), we can design the controller $\mathcal{F}$ for the system: 

\begin{equation}\label{eq:F_design}
\begin{array}{rl}
   \mathcal{F}^U: &  \bm{u}^U_k = \bm{K}^U_p \bm{e}^U_E + \bm{K}^U_d {\bm{\dot e}^U_E} +\prescript{\mathcal{C}}{\mathcal{B}}{\bm{T}}^U  \prescript{\mathcal{C}}{}{\bm{u}}^U_k \\[5pt]
    \mathcal{F}^S: &  \bm{u}^S_k = \bm{K}^S_p \bm{e}^S_E + \bm{K}^S_d {\bm{\dot e}^S_E} +\prescript{\mathcal{C}}{B}{\bm{T}}^S \prescript{\mathcal{C}}{}{\bm{u}}^S_k 
\end{array},
\end{equation}
above equation achieves that both robots simultaneously act as leader and follower through the use of the function $\Xi$, while executing their respective sub-tasks $E^U$ and $E^S$. VET simulates the tether effects on both robots to maintain the leader-follower formation. Unlike the approach in \cite{Yao2023},  {this formulation does not require a role-switching process when aiming to maintain LoS,} even when one robot is affected by perturbations. 



\section{Simulation Experiments}

\begin{figure}[tpb]
      \centering
       \includegraphics[width=0.5\textwidth]{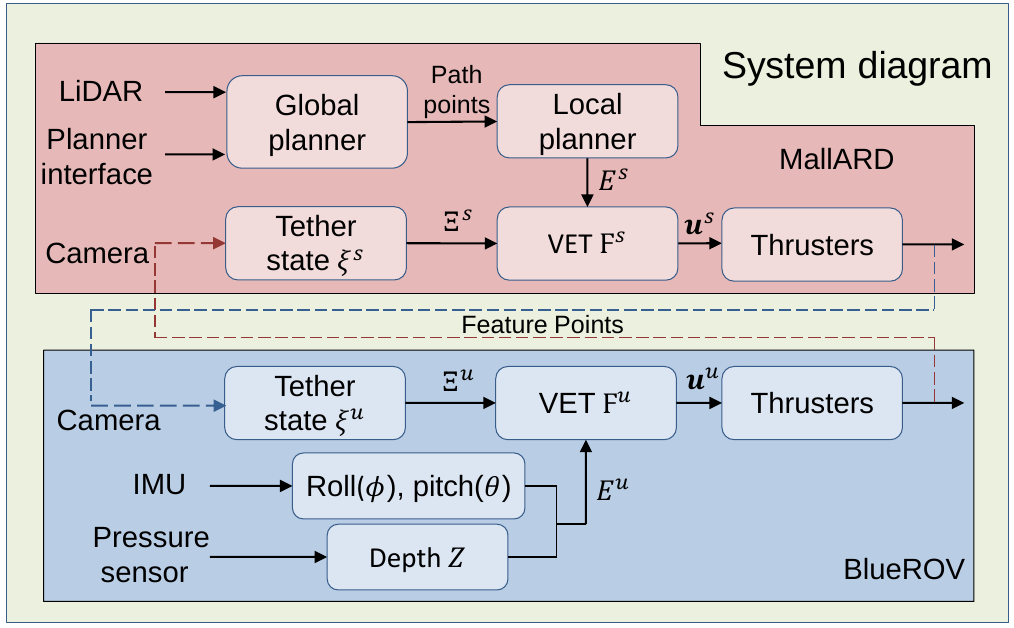}
      \caption{Diagram of the proposed system. Sensor inputs and planner interface are identical between the simulation and physical platform. Feature points are provided by the Apriltag library \cite{Wang2016}.
      }
      \label{fig:system_diagram}
  \end{figure}

To validate the performance of VET, we conducted a series of experiments in a simulated environment, CAVES-sim, featuring the BlueROV2 AUV and MallARD ASV. The simulations ran in a ROS (Robot Operating System) Melodic environment, built on Gazebo-11, executed on a PC equipped with an Intel(R) i7-10870H processor and 16GB RAM. We considered the hydrodynamics of both robots, simulating their operations in a 4.88\,m\,$\times$\,3.66\,m\,$\times$\,2.66\,m water tank. The setup also accounted for a simulated camera with a resolution of 640\,$\times$\,480\,px, tag detection, local planner \cite{Groves2019} for both robots, and global planner for MallARD. The system diagram is given in Fig.~\ref{fig:system_diagram}.

\begin{figure}[tpb]
      \centering
       \includegraphics[width=0.48\textwidth]{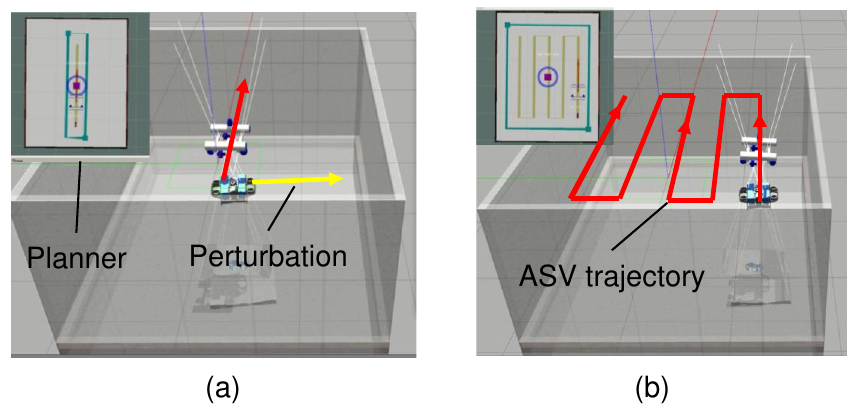}
      \caption{Gazebo simulation environment and planner's interface of CAVES-sim.
         (a) Schematic diagram of the perturbation experiment. The red arrow represents MallARD's planned path, while the yellow arrow indicates the perturbation.
        (b) Simulation of the navigation experiment, where the ASV's trajectory is depicted by red line arrows.
      }
      \label{fig:dis_full_sim}
  \end{figure}

Both linear and yaw-angular velocities as constrained as follows $\left[u^U_{max|x},u^U_{max|y}\right]^\top=\left[u^S_{max|x},u^S_{max|y}\right]^\top=[0.1\,m/s,0.1\,m/s]^\top$, and the maximum yaw-angular velocity as $u^U_{max|\psi}=u^S_{max|\psi}=0.2\,rad/s$. The desired tether state was set to $\xi_{d} = 0$, whilst the desired roll, pitch angle and depth for BlueROV were set to $[\phi^U_{d},\theta^U_{d}, Z^U_{d}]^\top = [0,0,-1\,m]^\top$. The tuning parameters of VET in function $\Xi$ are set to $[K_{s|p},K_{e|p},K_{e|d}]^\top = [0.5, 1, 0.15]^\top$. The subtask parameters of BlueROV and MallARD are set to $[\bm{K}^U_p, \bm{K}^U_d]^\top = [0.5, 0.15]^\top \bm{I}$, and $[\bm{K}^S_p, \bm{K}^S_d]^\top = [5, 5]^\top \bm{I}$, where $\bm{I}$ is a 6$\times$6 identity matrix.
Simulation experiments are organised into the following three sets:
\begin{enumerate}
    \item validate the system's stability in nominal scenario,
    \item evaluate perturbation handling (shown in Fig.~\ref{fig:dis_full_sim}.a), and
    \item assess navigation capabilities (shown in Fig.~\ref{fig:dis_full_sim}.b).
\end{enumerate}

\subsection{Experiment 1: Nominal setpoint control}
\subsubsection{Objective \& setup}
The primary goal of this experiment is to validate the convergence of Eq.~(\ref{eq:F_design}) for the proposed method. The positions of both robots are initialized as $\bm{x}^U_0 = [0, 0, -1, 0, 0, 0]^\top$ and $\bm{x}^S_0 = [0, 0, 0]^\top$. The BlueROV was then guided to a new position, $\bm{x}^U_{d} = [1, 1, -1, 0, 0, -1.57]^\top$, using the MallARD.

  \begin{figure}[tpb]
      \centering
       \includegraphics[width=0.48\textwidth]{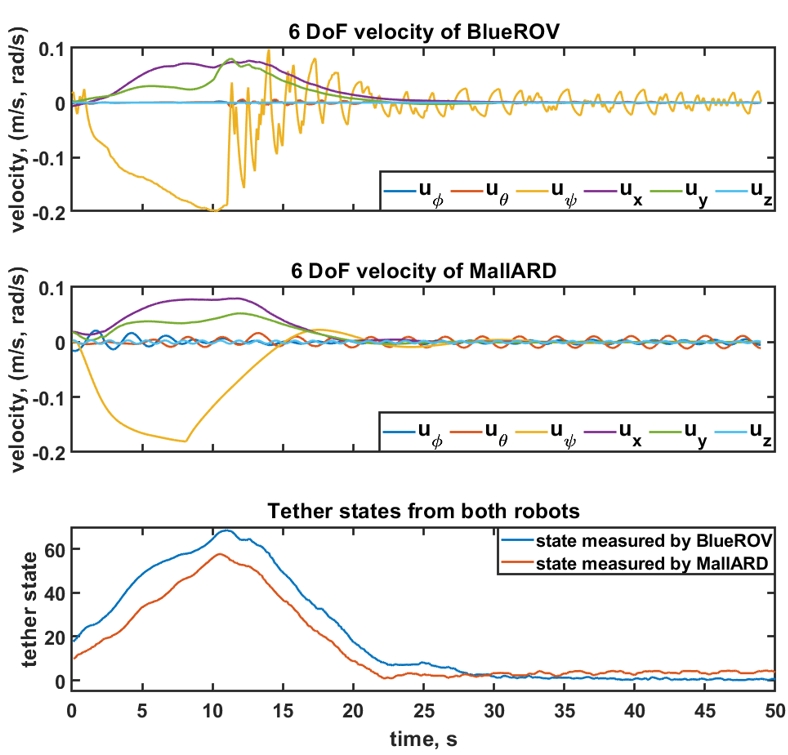}
      \caption{In the results of the convergence experiment, the time-varying 6DoF velocities of both robots are displayed in the top two figures. The bottom figure shows the measured tether states (as per (\ref{eq:system_target}), with $\xi_{d} = 0$) relative to each robot. Both robots' velocities and relative tether states converge to values close to zero, demonstrating that the proposed system is controllable in aquatic environments in a nominal scenario.
      }
      \label{fig:convergence}
  \end{figure}

\subsubsection{Result \& discussion}
Result of the convergence experiment is shown in Fig.~\ref{fig:convergence}, which presents the 6-DoF velocities and the relative states of both robots over time. Upon sending the desired position at $t = 0\,s$, the velocity of MallARD, represented as $[u^S_x, u^S_y, u^S_\psi]^\top$, increases initially and then converges to zero by $t = 16\,s$. Subsequently, the velocity of the BlueROV, denoted as $[u^U_x, u^U_y, u^U_\psi]^\top$, stabilises by $t = 20\,s$. The relative tether states, in our implementation, denoted by $|\bm{C}_{tag} - \bm{I}|$, as described in section \MakeUppercase{\romannumeral 5}, also stablise by $t=20\,s$. These observations validate the efficacy of Eq.~(\ref{eq:F_design}). Note that the values $[u_z, u_\phi, u_\theta]^\top$ exhibit some fluctuation beyond $t = 20\,s$, which is attributable to the effects of hydrodynamics.

\subsection{Experiment 2: Perturbations}
\subsubsection{Objective \& setup}
The objective of this experiment is to evaluate the system's resilience to external perturbations. We set the initial positions of both robots as $\bm{x}^U_0 = [0.1, 0, -1, 0, 0, 0]^\top$ and $\bm{x}^S_0 = [0.1, 0, 0]^\top$. With MallARD's guidance, the BlueROV moved towards a designated target $\bm{x}^U_{d}=[0.1, 3, -1, 0, 0, 0]^\top$. During their movement, a simulated external force perturbed the BlueROV, affecting both its trajectory and the formation with MallARD. As shown in Fig.~\ref{fig:dis_full_sim}.a, the yellow arrow pointing in the $-x$, originating at $y=0.75$, represents the perturbation.

  \begin{figure}[tpb]
      \centering
       \includegraphics[width=0.5\textwidth]{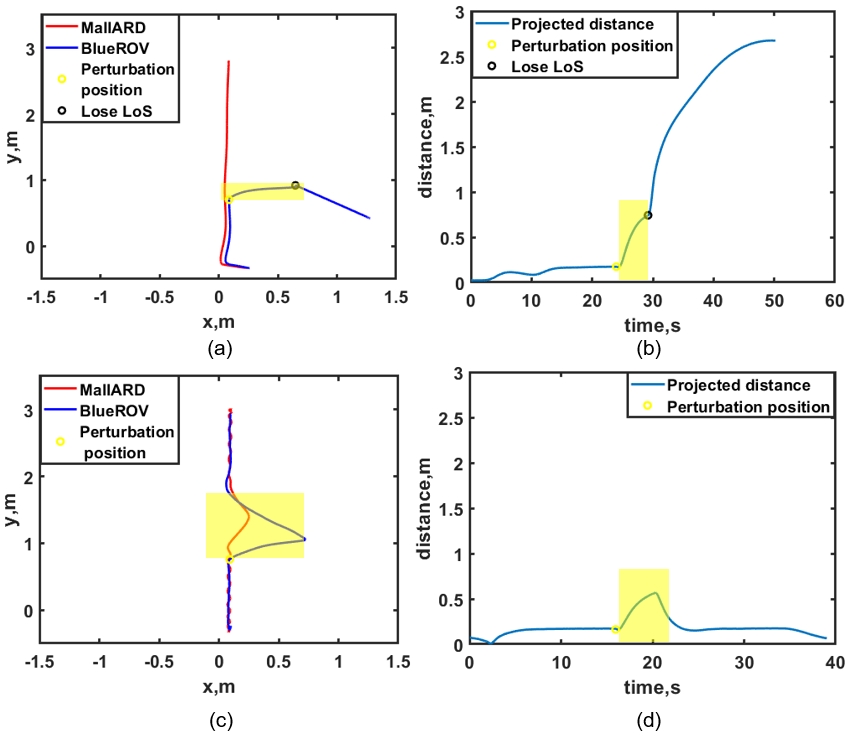}
      \caption{Projected trajectories and distance norm of BlueROV and MallARD on the \( Z_d^{U} = 0 \) plane during perturbation experiment without (top) and with (bottom) the VET-enhanced system. The yellow area highlights the regions affected by the perturbation. 
        .
      }
      \label{fig:dis_sim}
  \end{figure}

\subsubsection{Baseline}
In \cite{Yao2023}, the AUV and ASV maintain tracking when their projected distance norm remains below 0.6\,m. To highlight the effectiveness of the VET-enhanced formulation, we used the conventional IBVS-based AUV-ASV leader-follower formation from \cite{Yao2023} as our experimental baseline.

\subsubsection{Result \& discussion}
The results of the perturbation handling experiment are detailed in Fig.~\ref{fig:dis_sim}. The trajectories of BlueROV and MallARD are both acquired by Gazebo ground truth plugin\footnote{\url{https://classic.gazebosim.org/tutorials?tut=ros_gzplugins}}. Specifically, Fig.~\ref{fig:dis_sim}.a and Fig.~\ref{fig:dis_sim}.c depict the trajectories of the BlueROV and MallARD. The projected distance norms of the robots on the \( Z_d^{U} = 0 \) plane are presented in Fig.~\ref{fig:dis_sim}.b and Fig.~\ref{fig:dis_sim}.d, respectively. Notably, following the perturbation, the BlueROV deviated from its planned trajectory. In the baseline method, MallARD did not adapt its trajectory (shown in Fig.~\ref{fig:dis_sim}.a) to maintain the LoS of both robots. Consequently, the BlueROV lost LoS with MallARD 3 seconds post-perturbation, causing the projected distance norm to exceed 0.6\,m and the breaking of the formation. In contrast, in the VET-enhanced approach, MallARD proactively adjusted its path (as seen in Fig.~\ref{fig:dis_sim}.c) to maintain LoS with the BlueROV. As a result, the projected distance norm diminished (shown in Fig.~\ref{fig:dis_sim}.d) to an acceptable value (below 0.6\,m) within 5 seconds.  

    \begin{figure}[tpb]
      \centering
       \includegraphics[width=0.5\textwidth]{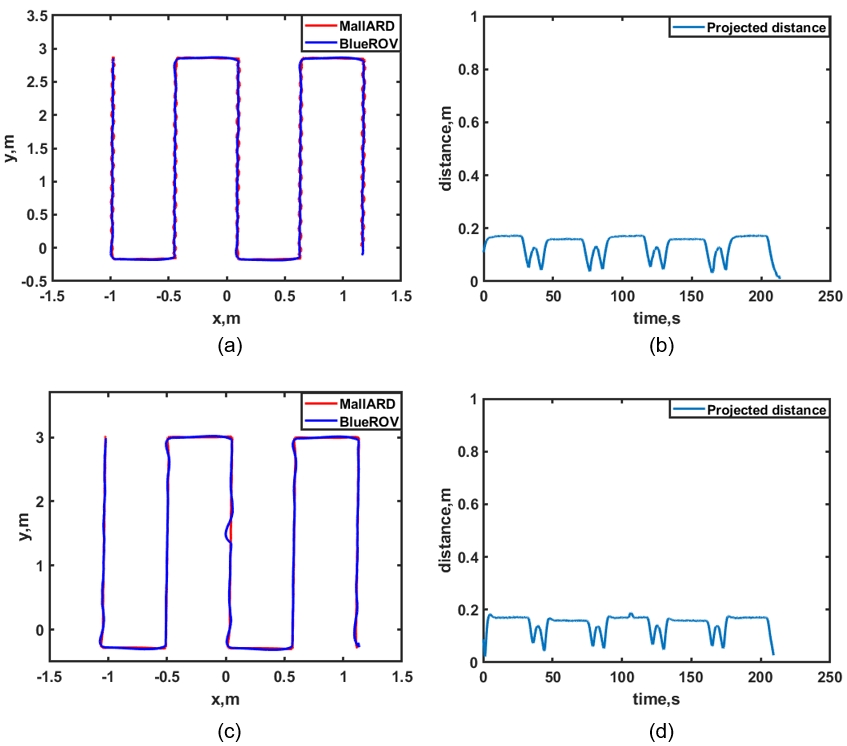}
      \caption{Lawnmower-patterned trajectories and distance norm of BlueROV and MallARD on the \( Z_d^{U} = 0 \) plane during navigation experiment. The top figure represents the baseline system while the bottom figure shows the results of the VET-enhanced system. 
      }
      \label{fig:full_sim}
  \end{figure}

\subsection{Experiment 3: Navigation}

\subsubsection{Objective \& setup}
This experiment is designed to evaluate the navigation capabilities of the BlueROV under the guidance of MallARD. Both vehicles start from their initial positions, with BlueROV at $\bm{x}^U_0 = [1.1, -0.25, -1, 0, 0, 0]^\top$  and MallARD at $\bm{x}^S_0 = [0, 0, 0]^\top$. They then follow a lawnmower-pattern path planned by the global planner \cite{groves2019m} to explore the whole tank. As shown in Fig.~\ref{fig:dis_full_sim}.b, the mission is to use MallARD to guide BlueROV through the path without any reliance on underwater localisation and communication techniques. 

\subsubsection{Baseline}
Similar to the perturbation experiment, the conventional IBVS-based AUV-ASV leader-follower formation was implemented as our experimental baseline.

\subsubsection{Result \& discussion}

Navigation results are presented in Fig.~\ref{fig:full_sim}.Under the guidance of MallARD, the BlueROV successfully navigated through the tank at its preset depth in both the VET-enhanced system and the baseline system. Furthermore, it maintained its formation with MallARD, with the maximum projected distance norm remaining below 0.2\,m. 
The performances of the VET-enhanced and baseline systems are largely comparable. This similarity arises because the Gazebo-based simulation does not replicate some key factors that can affect the system's LoS stability, such as varying ambient lighting, water distortion, and suspended particles in water. Hence, to further validate the effectiveness of VET in enhancing leader-follower formation under extreme environments, more realistic experiments should be conducted.

\section{Real tank experiments}

In addition to the simulation experiments in Section \MakeUppercase{\romannumeral 7}, we implemented the VET and CAVES on the physical platform using MallARD ASV and a customised BlueROV2 AUV. As with the simulation packages, MallARD ASV\footnote{\url{https://robotki.github.io/entries/robots-aquatic-mallard.html}} and the ROSify instructions of BlueROV2\footnote{\url{git@github.com:drunkbot/ROSIFY_bluerov.git}} have also been made open-source resources. The system diagram is identical to that in the simulation, given in Fig.~\ref{fig:system_diagram}.

  \begin{figure}[tpb]
      \centering
       \includegraphics[width=0.5\textwidth]{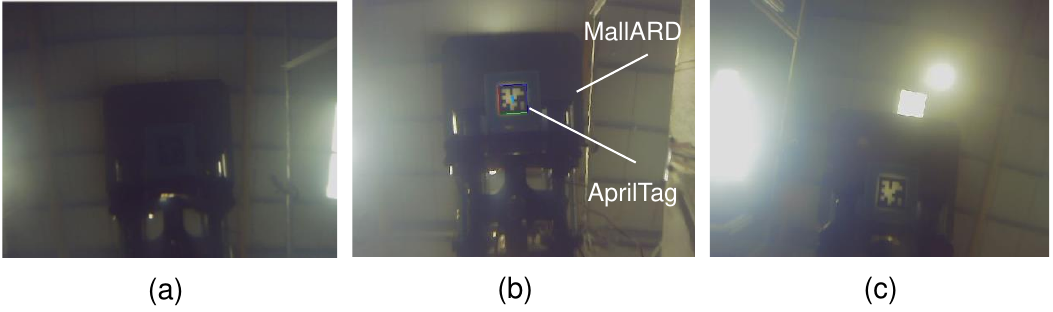}
      \caption{Camera perspective from the BlueROV directed upwards for marker detection.
        (a) ordinary marker on the water surface, undetected.
        (b) backlit marker on the surface, detected by the robot.
        (c) despite being backlit, the marker on the surface goes undetected occasionally due to ambient lighting. 
      }
      \label{fig:active_marker}
  \end{figure}

  \begin{figure*}[tpb]
      \centering
       \includegraphics[width=\textwidth]{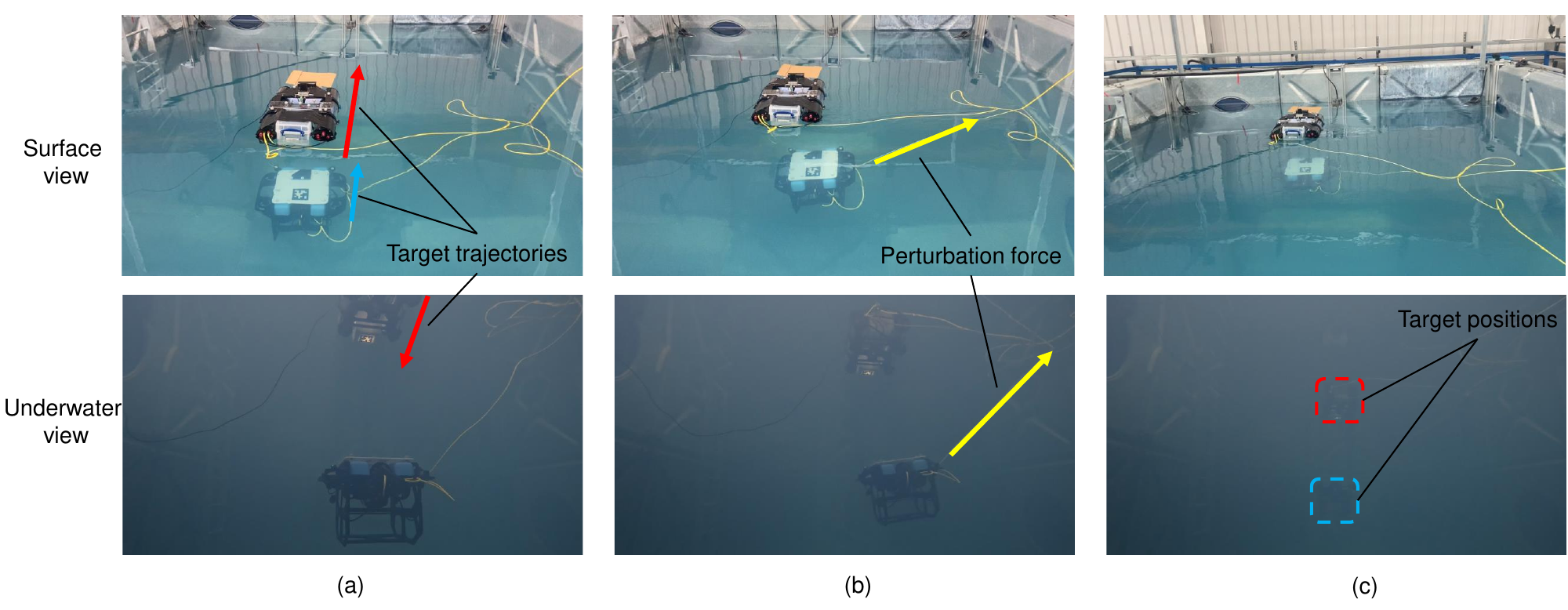}
      \caption{Surface view and sub-surface view of the perturbation experiment conducted in a 4.88\,m\,$\times$\,3.66\,m\,$\times$\,2.66\,m experimental tank. a), the starting position of BlueROV and MallARD, b), a perturbation attempting to disrupt the formation (LoS) of the robots, c), the target position of both robots. It is important to note that the tether on both robots is for debugging and online data viewing only, all packages are capable of running purely on the robots' onboard PC.
      }
      \label{fig:dis_fig}
  \end{figure*}

  \begin{figure*}[tpb]
      \centering
       \includegraphics[width=\textwidth]{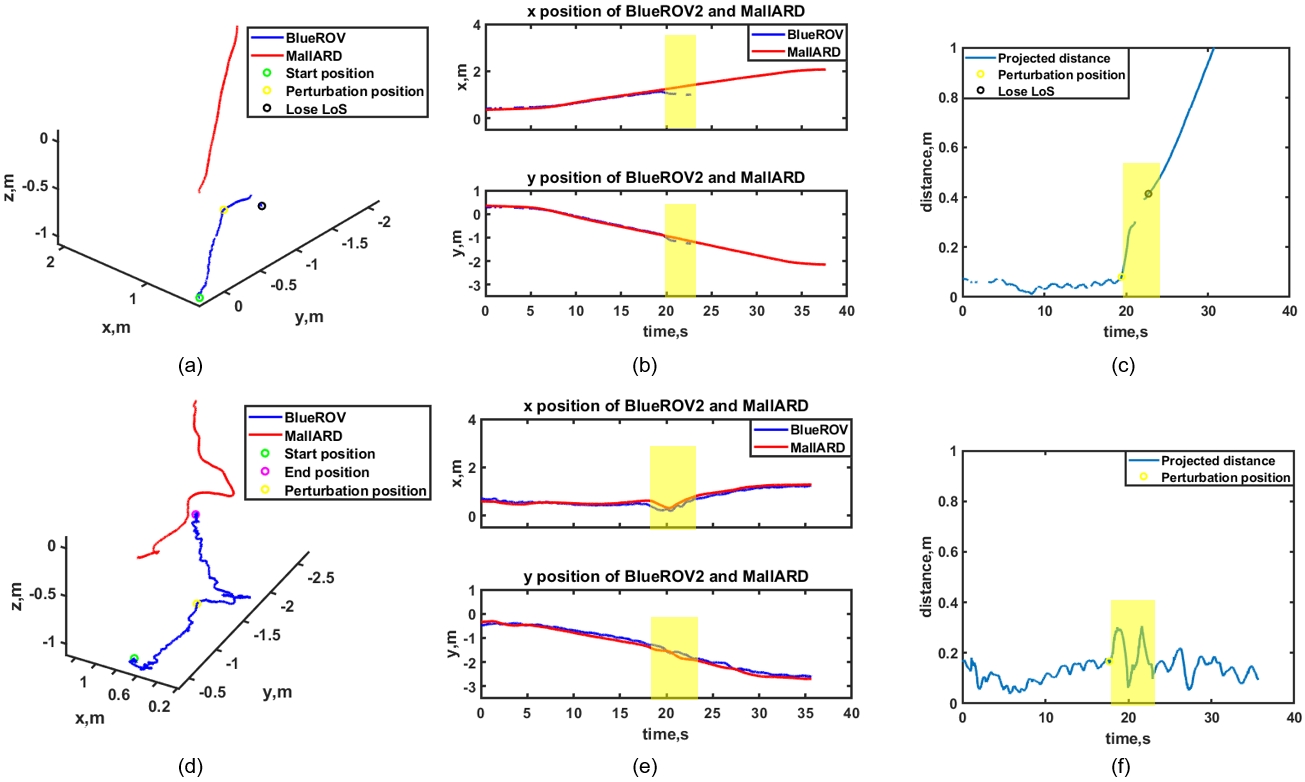}
      \caption{Result of perturbation experiment conducted in a tank. The yellow area highlights the regions affected by the perturbation. The top figure depicts the results for the baseline system, illustrating trajectories, $xy$-positions, and projected distances. Conversely, the bottom presents the results for the VET-enhanced system, similarly detailing trajectories, $xy$-positions, and projected distances, respectively.  Mission shown in Fig.~\ref{fig:dis_fig}.
      }
      \label{fig:dis_exp}
  \end{figure*}

\subsection{Experiment Setup}
As detailed in the prior section, the VET-enhanced system, through simulation, showcased an advantage over the traditional IBVS formulation, especially in managing perturbations. However, certain critical aspects that might influence the system's LoS stability are not replicable in the Gazebo simulation environment. To further validate the effectiveness of the VET formulation, two sets of physical experiments focused on perturbation and navigation are conducted and evaluated in this section. 

The linear and yaw-angular velocities as constrained same as that in the simulation, $\left[u^U_{max|x},u^U_{max|y}\right]^\top=\left[u^S_{max|x},u^S_{max|y}\right]^\top=[0.1\,m/s,0.1\,m/s]^\top$, and the maximum yaw-angular velocity as $u^U_{max|\psi}=u^S_{max|\psi}=0.2\,rad/s$. The desired tether state was set to $\xi_{d} = 0$, the desired roll, pitch angle, and depth for BlueROV was set to $[\phi^U_{d},\theta^U_{d}, Z^U_{d}]^\top = [0,0,-1\,m]^\top$.
The tuning parameters of VET in function $\Xi$ are set to $[K_{s|p},K_{e|p},K_{e|d}]^\top = [0.35, 1, 0.15]^\top$. The subtask parameters of BlueROV and MallARD are set to $[\bm{K}^U_p, \bm{K}^U_d]^\top = [0.5, 0.15]^\top \bm{I}$, and $[\bm{K}^S_p, \bm{K}^S_d]^\top = [1, 0.5]^\top \bm{I}$, where $\bm{I}$ is a 6$\times$6 identity matrix.

 \subsubsection{Physical Platform}
Both robots utilize a low-cost Raspberry Pi 4 as their onboard PC, with ROS Melodic serving as the middleware. The PX4-based flight computing unit (FCU) \cite{Meier2015} manages the low-level control of actuators. Additionally, MallARD is outfitted with a SICK 2D LiDAR TiM5xx and employs the hector SLAM with millimeter-level accuracy \cite{groves2019m}. The cameras on both robots are low-cost Raspberry Pi USB cameras, offering a resolution of 640\,$\times$\,480\,px.

\subsubsection{Baseline} 
Consistent with the simulation experiments, we adopted the system proposed in \cite{Yao2023}, where the BlueROV follows MallARD. This serves as the experimental baseline for both the perturbation and navigation experiments.

  \begin{figure*}[tpb]
      \centering
       \includegraphics[width=\textwidth]{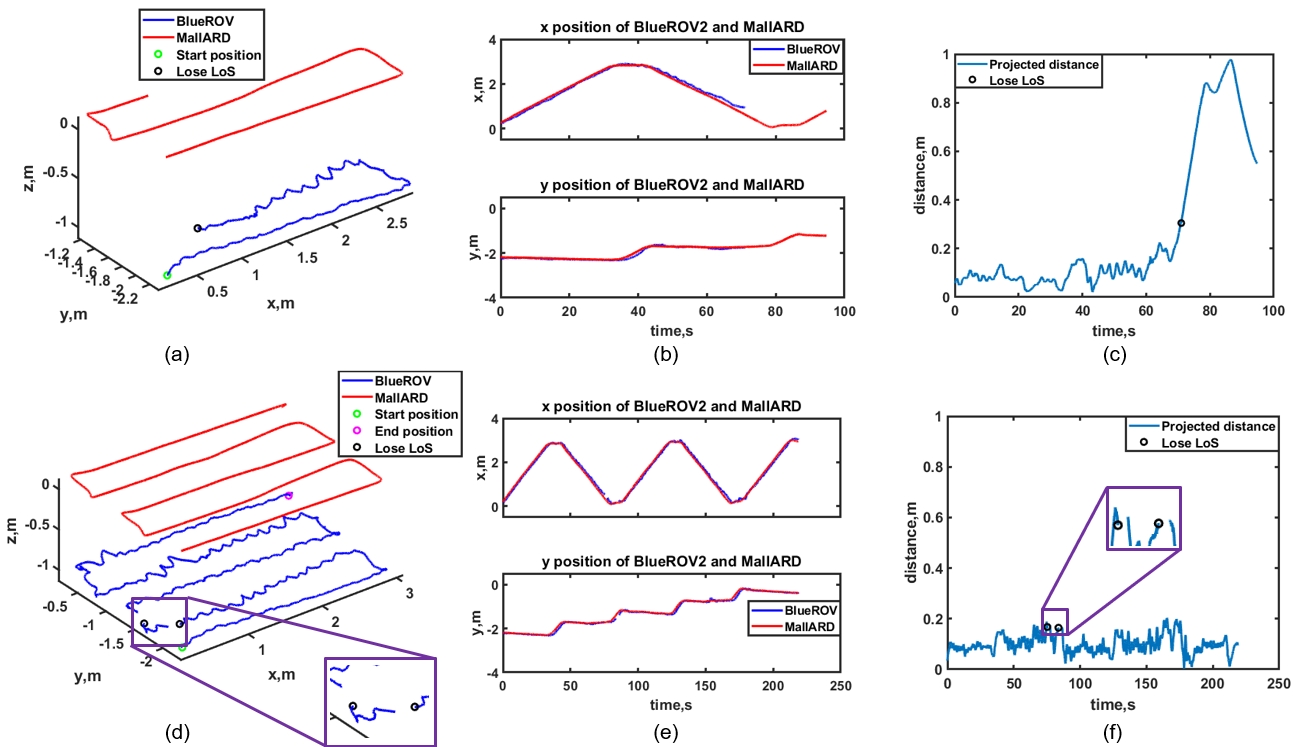}
      \caption{ Result of navigation experiment conducted in an experimental tank.  The top figure depicts the results of the baseline system, illustrating trajectories, $xy$-positions, and projected distances. Conversely, the bottom figure presents the results for the VET-enhanced system, similarly detailing trajectories, $xy$-positions, and projected distances, respectively. Despite the loss of tag detections, as highlighted in d) and f), robots in the VET-enhanced system remain connected and finish the mission. Mission shown in Fig.~\ref{fig:caves_display}.
      }
      \label{fig:full_exp}
  \end{figure*}

  \begin{figure}[tpb]
      \centering
       \includegraphics[width=0.48\textwidth]{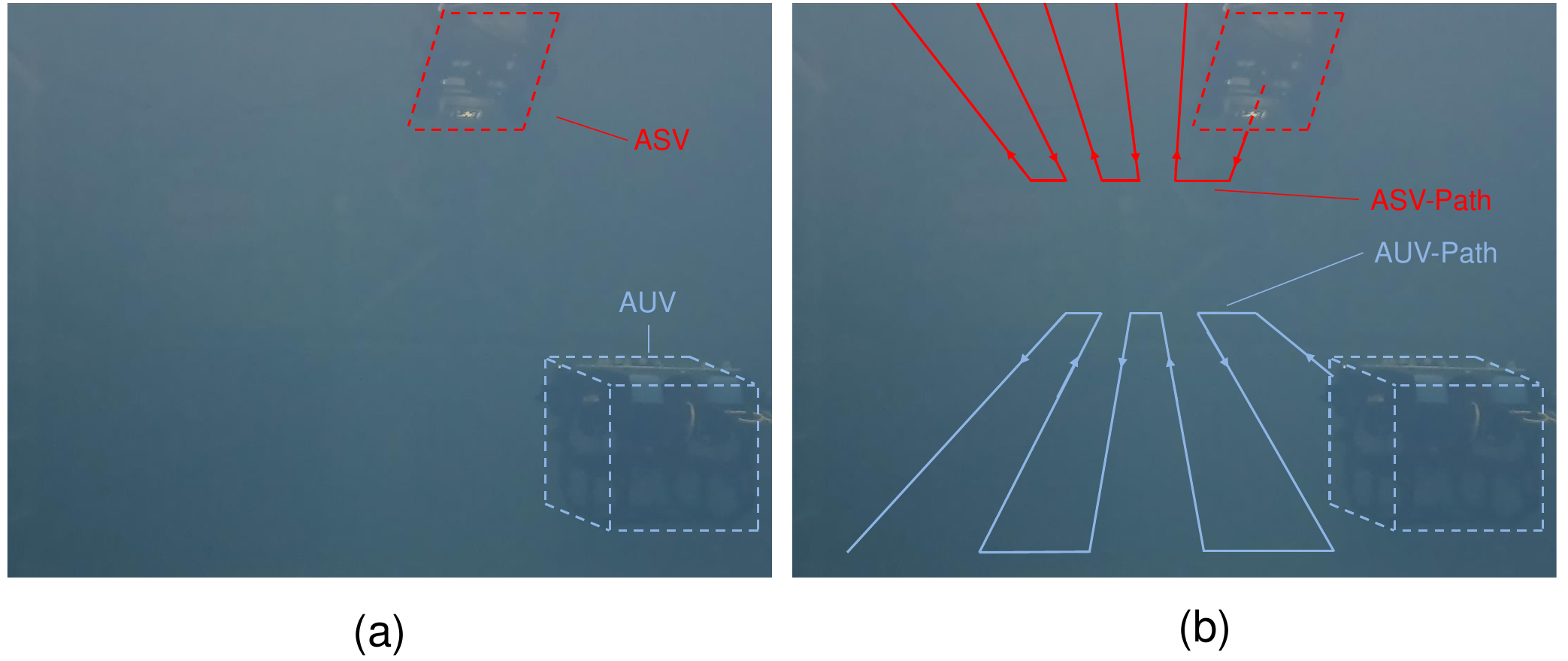}
      \caption{CAVES in turbid experimental pond, RAICo1, West Cumbria, England. Robots are highlighted in blue and red dashed lines. a) displays the conditions of the environment, poor lighting conditions, lack of dense and salient features that prohibit effective application of visual-based SLAM algorithms, and no access to external auxiliary facilities; b) shows the aim of the experiment, to achieve robust autonomous navigation of the AUV following the displayed path under the guidance of the ASV.
      }
      \label{fig:caves_display}
  \end{figure}

\subsubsection{Active fiducial marker}

Unlike the approach in \cite{Yao2023}, where the follower's camera is always oriented downwards (restricting the follower to remain above the leader) due to the upward orientation causing tag detection loss from ambient lighting (as depicted in Fig.~\ref{fig:active_marker}.a). In our experiments, we employ a unique backlit marker to address this issue, illustrated in Fig.~\ref{fig:active_marker}.b, our design successfully allows an underwater robot to track a surface robot using an upward-facing camera. However, it's worth noting that even with backlighting, intense overhead illumination can still lead to tag detection loss, as illustrated in Fig.~\ref{fig:active_marker}.c. The implications of this scenario on the system will be discussed in the results section.

\subsubsection{Ground truth} 
The trajectory of MallARD was acquired using the on-board 2D SLAM from \cite{Groves2019}. However, due to the low visibility in the experimental tank, attributed to its water quality, the general infrastructure-based underwater ground truth system such as Qualisys does not function as it should. Nevertheless, the presence of tag-based feature points enables us to compute the BlueROV2's trajectory relative to MallARD's self-localisation output using a Perspective-n-Point (PnP) approach, 

\begin{equation}
   \prescript{\mathcal{W}}{}{\bm{x}}^U = \prescript{\mathcal{W}}{\mathcal{B}}{\bm{T}}^S \prescript{\mathcal{B}}{\mathcal{C}}{\bm{T}}^S \prescript{\mathcal{C}}{\mathcal{T}}{\bm{T}}^{S\rightarrow U}
\end{equation}
where $\prescript{\mathcal{W}}{}{\bm{x}}^U$ denotes the position of BlueROV in world-fixed frame $\mathcal{W}$, transformation $\prescript{\mathcal{W}}{\mathcal{B}}{\bm{T}}^S$ and $\prescript{\mathcal{B}}{\mathcal{C}}{\bm{T}}^S$ represent the world-fixed frame $\mathcal{W}$ to body frame $\mathcal{B}$ and body frame $\mathcal{B}$ to camera frame $\mathcal{C}$ of MallARD, respectively. $\prescript{\mathcal{C}}{\mathcal{T}}{\bm{T}}^{S \rightarrow U}$ denotes the transformation from the camera frame $\mathcal{C}^S$ of MallARD to the tag frame $\mathcal{T}^U$ of BlueROV. In our experiments, $\prescript{\mathcal{C}}{\mathcal{T}}{\bm{T}}^{S \rightarrow U}$ was provided by Apriltag \cite{Wang2016} library, $\prescript{\mathcal{W}}{}{\bm{x}}^U$ was computed by post-process of the recorded data, which is available at \footnote{\url{https://livemanchesterac-my.sharepoint.com/:f:/g/personal/kanzhong_yao_postgrad_manchester_ac_uk/Ejve1Y7Rsf5JkHqf8olLfWoBR-y_uwPZIknewox8QQ9hPg?e=7N63J1}}.

\subsection{Experiment 1: Perturbations}
\subsubsection{Setup} 
Consistent with that in experiment 2 of the simulation, the positions of both robots are initialized as $\bm{x}^U_0 = [0.2, -0.5, -1, 0, 0, 0]^\top$ and $\bm{x}^S_0 = [0.2, -0.5, 0]^\top$. Under the guidance of MallARD, the BlueROV moved towards a designated target $\bm{x}^U_{d}=[1, -2.5, -1, 0, 0, 0]^\top$. During their movements, the operator pulled the tether attached to the right side of BlueROV as an external perturbation, affecting its trajectory and the formation with MallARD. As shown in Fig.~\ref{fig:dis_fig}.a, the yellow arrow pointing in the $-x$, originating at $y=-1$, represents the perturbation.

\subsubsection{Result \& discussion}
The results are detailed in Fig.~\ref{fig:dis_exp}. Specifically, a), b), d), and e) depict the trajectories and $xy$-positions of the BlueROV and MallARD during the experiment. While c) and f) depict the projected distances of the robots on the \( Z_d^{U} = 0 \) plane. Notably, following the perturbation, the BlueROV deviated from its original trajectory. In the baseline method, MallARD did not adapt its trajectory (shown in Fig.~\ref{fig:dis_exp}.a) and b). Consequently, the BlueROV lost LoS with MallARD 4 seconds after the perturbation, ((shown in Fig.~\ref{fig:dis_exp}.c)) causing the projected distance to exceed 0.3\,m and the breaking of formation. In contrast, in the VET-enhanced system, MallARD proactively adjusted its path (as seen in Fig.~\ref{fig:dis_exp}.d) and e), to maintain LoS with the BlueROV. As a result, the projected distance diminished (shown in Fig.~\ref{fig:dis_exp}.f) to an acceptable value (below 0.3\,m) within 5 seconds.

\subsection{Experiment 2: Navigation}

\subsubsection{Setup}
Consistent with that in experiment 3 of the simulation, both vehicles start from their initial positions, with BlueROV at $\bm{x}^U_0 = [0.2, -2.2, -1, 0, 0, 0]^\top$  and MallARD at $\bm{x}^S_0 = [0.2, -2.2, 0]^\top$. They then follow a lawnmower-pattern path planned by the global planner \cite{groves2019m} to explore the whole tank. As shown in Fig.~\ref{fig:caves_display}, the mission is to use MallARD to guide BlueROV through the path without any reliance on underwater localisation and communication techniques.

\subsubsection{Result \& discussion}
    
Navigation results are presented in Fig.~\ref{fig:full_exp}. Unlike that in the simulation, when the intense overhead illumination caused the loss of tag detection, as shown in Fig.~\ref{fig:active_marker}.c), which happened at $t=71$\,s during the navigation experiment, the formation of the baseline method was disrupted (Fig.~\ref{fig:full_exp}.a) and b)). Following such disruption, the projected distance between the robots then exceeds 0.6\,m and the mission failed (Fig.~\ref{fig:full_exp}.c)). However, in the VET-enhanced approach, BlueROV successfully navigated through the tank at its preset depth under the guidance of MallARD, despite the loss of tag detections at $t=75$\,s and $t=85$\,s, as illustrated in Fig.~\ref{fig:full_exp}.d) and f). Therefore, it is reasonable to say that the VET-enhanced approach outperforms the conventional IBVS-based method in terms of multi-agent navigation in extreme aquatic environments. 

\subsection{Summary}
From our results, it is noteworthy that even though certain aspects of real aquatic environments cannot be precisely replicated in the Gazebo environment—leading to different results in navigation experiments—there is a notable consistency in magnitude between CAVES-sim and the physical CAVES in terms of stable projected distances, recovery time from perturbations, and navigation durations. 
Such consistency validates the accuracy of our simulation and underscores the adaptability of the VET-enhanced system when transitioning from a simulation to real-world scenarios.
Given the rigorous testing conditions necessary for underwater robot development, CAVES-sim emerges as a reliable, open-sourced platform for initial phases of development and preliminary algorithm validation.
\section{CONCLUSION \& FUTURE WORK}

In this article, the problem of underwater navigation under incomplete state measurement in extreme aquatic environments was defined. To address this problem, a new approach called Virtual Elastic Tether (VET) was proposed. First, the underwater navigation problem was re-formulated into a multi-agent leader-follower problem, then the generalised formulation of the VET was conceptualised and modeled. To validate such an approach, a sim-to-real Cooperative Aquatic Vehicle Exploration System (CAVES) was proposed. Based on CAVES, a visual AUV-ASV leader-follower formation implementation of VET was formulated. Finally, the controllability, perturbation handling capability, and navigation capability were validated, both in simulation and on the physical platform. An IBVS-based approach was also implemented as a baseline system in the sim-to-real validation, results indicate that the VET-enhanced system outperforms the conventional approach in terms of perturbation handling and navigation under extreme aquatic environments, proving the feasibility of deploying such a system in real underwater environments. 


It must be noted that, while the proposed strategy is developed, specifically, for a multi-agent underwater system, it can be adapted to other platforms such as aerial and ground-based robots. {The VET approach is designed to offer a control architecture that enables users to define customised parameters for various systems. While the stability of the current system has been validated experimentally, it has not yet been substantiated through a general and theoretical framework. Therefore, future work will focus on applying the VET approach to a broader range of platforms and examining its effectiveness in contexts beyond underwater applications.} Furthermore, during the formulation of VET, we observed that such inter-robotic connections might offer advantages in multi-agent obstacle avoidance during formation missions. This potential benefit will also be a focal point of our upcoming investigations.

\section*{Acknowledgments}
We thank Jessica Paterson and Dr. Melissa Sandison, for their help with the experimental setup.  We sincerely appreciate the valuable advice that Dr. Wei Cheah and Dr. Bruno Adorno gave regarding the original draft.

\bibliographystyle{IEEEtran}
\bibliography{reference.bib}

\begin{thebibliography}{10}
\providecommand{\url}[1]{#1}
\csname url@samestyle\endcsname
\providecommand{\newblock}{\relax}
\providecommand{\bibinfo}[2]{#2}
\providecommand{\BIBentrySTDinterwordspacing}{\spaceskip=0pt\relax}
\providecommand{\BIBentryALTinterwordstretchfactor}{4}
\providecommand{\BIBentryALTinterwordspacing}{\spaceskip=\fontdimen2\font plus
\BIBentryALTinterwordstretchfactor\fontdimen3\font minus
  \fontdimen4\font\relax}
\providecommand{\BIBforeignlanguage}[2]{{%
\expandafter\ifx\csname l@#1\endcsname\relax
\typeout{** WARNING: IEEEtran.bst: No hyphenation pattern has been}%
\typeout{** loaded for the language `#1'. Using the pattern for}%
\typeout{** the default language instead.}%
\else
\language=\csname l@#1\endcsname
\fi
#2}}
\providecommand{\BIBdecl}{\relax}
\BIBdecl

\bibitem{legacy_pond}
\BIBentryALTinterwordspacing
{Office for Nuclear Regulation}, ``{Divers enter legacy pond at Sellafield},''
  2023, [Online; accessed 8-July-2023]. [Online]. Available:
  \url{https://news.onr.org.uk/2023/03/divers-enter-legacy-pond-at-sellafield/}
\BIBentrySTDinterwordspacing

\bibitem{fuel_pond}
\BIBentryALTinterwordspacing
{JAMES TEMPERTON}, ``{Inside Sellafield: how the UK's most dangerous nuclear
  site is cleaning up its act},'' 2016, [Online; accessed 8-July-2023].
  [Online]. Available:
  \url{https://www.wired.co.uk/article/inside-sellafield-nuclear-waste-decommissioning}
\BIBentrySTDinterwordspacing

\bibitem{Watson2020}
S.~Watson, D.~A. Duecker, and K.~Groves, ``Localisation of unmanned underwater
  vehicles ({UUVs}) in complex and confined environments: A review,''
  \emph{Sensors}, vol.~20, no.~21, p. 6203, Oct. 2020.

\bibitem{Zhang2023}
B.~Zhang, D.~Ji, S.~Liu, X.~Zhu, and W.~Xu, ``Autonomous underwater vehicle
  navigation: A review,'' \emph{Ocean Engineering}, vol. 273, p. 113861, apr
  2023.

\bibitem{oikawa2016r}
K.~Oikawa, ``R\&d on robots for the decommissioning of fukushima daiichi nps,''
  \emph{Int. Res. Inst. Nucl. Decommissioning, Tokyo, Japan, Tech. Rep}, 2016.

\bibitem{Zhou2022}
X.~Zhou, X.~Wen, Z.~Wang, Y.~Gao, H.~Li, Q.~Wang, T.~Yang, H.~Lu, Y.~Cao,
  C.~Xu, and F.~Gao, ``Swarm of micro flying robots in the wild,''
  \emph{Science Robotics}, vol.~7, no.~66, May 2022.

\bibitem{Bird2021}
B.~Bird, M.~Nancekievill, A.~West, J.~Hayman, C.~Ballard, W.~Jones, S.~Ross,
  T.~Wild, T.~Scott, and B.~Lennox, ``Vega{\textemdash}a small, low cost,
  ground robot for nuclear decommissioning,'' \emph{Journal of Field Robotics},
  vol.~39, no.~3, pp. 232--245, nov 2021.

\bibitem{Jiang2022}
Z.~Jiang, F.~Jovan, P.~Moradi, T.~Richardson, S.~Bernardini, S.~Watson,
  A.~Weightman, and D.~Hine, ``A multirobot system for autonomous deployment
  and recovery of a blade crawler for operations and maintenance of offshore
  wind turbine blades,'' \emph{Journal of Field Robotics}, vol.~40, no.~1, pp.
  73--93, sep 2022.

\bibitem{GonzalezGarcia2020}
J.~Gonz{\'{a}}lez-Garc{\'{\i}}a, A.~G{\'{o}}mez-Espinosa, E.~Cuan-Urquizo,
  L.~G. Garc{\'{\i}}a-Valdovinos, T.~Salgado-Jim{\'{e}}nez, and J.~A.~E.
  Cabello, ``Autonomous underwater vehicles: Localization, navigation, and
  communication for collaborative missions,'' \emph{Applied Sciences}, vol.~10,
  no.~4, p. 1256, feb 2020.

\bibitem{Qu2016}
F.~Qu, Z.~Wang, L.~Yang, and Z.~Wu, ``A journey toward modeling and resolving
  doppler in underwater acoustic communications,'' \emph{{IEEE} Communications
  Magazine}, vol.~54, no.~2, pp. 49--55, Feb. 2016.

\bibitem{Lategahn2011}
H.~Lategahn, A.~Geiger, and B.~Kitt, ``Visual slam for autonomous ground
  vehicles,'' in \emph{2011 IEEE International Conference on Robotics and
  Automation}.\hskip 1em plus 0.5em minus 0.4em\relax IEEE, May 2011.

\bibitem{Duecker_2020}
D.~A. Duecker, F.~Steinmetz, E.~Kreuzer, and C.~Renner, ``Micro {AUV}
  localization for agile navigation with low-cost acoustic modems,'' in
  \emph{2020 {IEEE}/{OES} Autonomous Underwater Vehicles Symposium
  ({AUV})(50043)}.\hskip 1em plus 0.5em minus 0.4em\relax {IEEE}, sep 2020.

\bibitem{Yang_2017}
M.~Yang, X.~Li, Y.~Yang, and X.~Meng, ``Characteristic analysis of underwater
  acoustic scattering echoes in the wavelet transform domain,'' \emph{Journal
  of Marine Science and Application}, vol.~16, no.~1, pp. 93--101, jan 2017.

\bibitem{Duecker2020b}
D.~A. Duecker, N.~Bauschmann, T.~Hansen, E.~Kreuzer, and R.~Seifried, ``Towards
  micro robot hydrobatics: Vision-based guidance, navigation, and control for
  agile underwater vehicles in confined environments,'' in \emph{2020 IEEE/RSJ
  International Conference on Intelligent Robots and Systems (IROS)}.\hskip 1em
  plus 0.5em minus 0.4em\relax IEEE, Oct 2020.

\bibitem{Tolba2017}
S.~Tolba and R.~Ammar, ``Virtual tether search: A self-constraining search
  algorithm for swarms in an open ocean,'' in \emph{2017 {IEEE} Symposium on
  Computers and Communications ({ISCC})}.\hskip 1em plus 0.5em minus
  0.4em\relax {IEEE}, jul 2017.

\bibitem{Wiech2018}
J.~Wiech, V.~A. Eremeyev, and I.~Giorgio, ``Virtual spring damper method for
  nonholonomic robotic swarm self-organization and leader following,''
  \emph{Continuum Mechanics and Thermodynamics}, vol.~30, no.~5, pp.
  1091--1102, apr 2018.

\bibitem{Yao2023}
K.~Yao, N.~Bauschmann, T.~L. Alff, W.~Cheah, D.~A. Duecker, K.~Groves,
  O.~Marjanovic, and S.~Watson, ``Image-based visual servoing switchable
  leader-follower control of heterogeneous multi-agent underwater robot
  system,'' in \emph{2023 IEEE International Conference on Robotics and
  Automation (ICRA)}, 2023, pp. 5200--5206.

\bibitem{Wang2023}
G.~Wang, J.~Qin, Q.~Liu, Q.~Ma, and C.~Zhang, ``Image-based visual servoing of
  quadrotors to arbitrary flight targets,'' \emph{{IEEE} Robotics and
  Automation Letters}, vol.~8, no.~4, pp. 2022--2029, apr 2023.

\bibitem{Lin2021}
J.~Lin, Z.~Miao, H.~Zhong, W.~Peng, Y.~Wang, and R.~Fierro, ``Adaptive
  image-based leader{\textendash}follower formation control of mobile robots
  with visibility constraints,'' \emph{{IEEE} Transactions on Industrial
  Electronics}, vol.~68, no.~7, pp. 6010--6019, Jul. 2021.

\bibitem{Bechlioulis2019a}
C.~P. Bechlioulis, S.~Heshmati-alamdari, G.~C. Karras, and K.~J. Kyriakopoulos,
  ``Robust image-based visual servoing with prescribed performance under field
  of view constraints,'' \emph{{IEEE} Transactions on Robotics}, vol.~35,
  no.~4, pp. 1063--1070, aug 2019.

\bibitem{Griffiths_2016}
A.~Griffiths, A.~Dikarev, P.~R. Green, B.~Lennox, X.~Poteau, and S.~Watson,
  ``{AVEXIS}{\textemdash}aqua vehicle explorer for in-situ sensing,''
  \emph{{IEEE} Robotics and Automation Letters}, vol.~1, no.~1, pp. 282--287,
  jan 2016.

\bibitem{Bhat2020}
S.~Bhat, I.~Torroba, O.~Ozkahraman, N.~Bore, C.~I. Sprague, Y.~Xie, I.~Stenius,
  J.~Severholt, C.~Ljung, J.~Folkesson, and P.~Ogren, ``A cyber-physical system
  for hydrobatic {AUVs}: System integration and field demonstration,'' in
  \emph{2020 {IEEE}/{OES} Autonomous Underwater Vehicles Symposium
  ({AUV})(50043)}.\hskip 1em plus 0.5em minus 0.4em\relax {IEEE}, sep 2020.

\bibitem{Edge2020a}
C.~Edge, S.~Sakib~Enan, M.~Fulton, J.~Hong, J.~Mo, K.~Barthelemy, H.~Bashaw,
  B.~Kallevig, C.~Knutson, K.~Orpen, and J.~Sattar, ``Design and experiments
  with loco auv: A low cost open-source autonomous underwater vehicle,'' in
  \emph{2020 IEEE/RSJ International Conference on Intelligent Robots and
  Systems (IROS)}.\hskip 1em plus 0.5em minus 0.4em\relax IEEE, Oct. 2020.

\bibitem{Chen2021}
J.~Chen, C.~Sun, and A.~Zhang, ``Autonomous navigation for adaptive unmanned
  underwater vehicles using fiducial markers,'' in \emph{2021 {IEEE}
  International Conference on Robotics and Automation ({ICRA})}.\hskip 1em plus
  0.5em minus 0.4em\relax {IEEE}, may 2021.

\bibitem{Skulstad2019}
R.~Skulstad, G.~Li, T.~I. Fossen, B.~Vik, and H.~Zhang, ``Dead reckoning of
  dynamically positioned ships: Using an efficient recurrent neural network,''
  \emph{{IEEE} Robotics $\&$ Automation Magazine}, vol.~26, no.~3, pp. 39--51,
  sep 2019.

\bibitem{Rahman2022}
S.~Rahman, A.~Q. Li, and I.~Rekleitis, ``{SVIn}2: A multi-sensor fusion-based
  underwater {SLAM} system,'' \emph{The International Journal of Robotics
  Research}, p. 027836492211102, jul 2022.

\bibitem{Xanthidis_2020}
M.~Xanthidis, N.~Karapetyan, H.~Damron, S.~Rahman, J.~Johnson,
  A.~O{\textquotesingle}Connell, J.~M. O{\textquotesingle}Kane, and
  I.~Rekleitis, ``Navigation in the presence of obstacles for an agile
  autonomous underwater vehicle,'' in \emph{2020 {IEEE} International
  Conference on Robotics and Automation ({ICRA})}.\hskip 1em plus 0.5em minus
  0.4em\relax {IEEE}, May 2020.

\bibitem{Xanthidis2023a}
M.~Xanthidis, E.~Kelasidi, and K.~Alexis, ``{ResiPlan}: Closing the
  planning-acting loop for safe underwater navigation,'' in \emph{2023 {IEEE}
  International Conference on Robotics and Automation ({ICRA})}.\hskip 1em plus
  0.5em minus 0.4em\relax {IEEE}, may 2023.

\bibitem{groves2019m}
K.~Groves, M.~Dimitrov, H.~Peel, O.~Marjanovic, and B.~Lennox, ``Model
  identification of a small omnidirectional aquatic surface vehicle: a
  practical implementation,'' in \emph{2020 IEEE/RSJ International Conference
  on Intelligent Robots and Systems (IROS)}, 2020, pp. 1813--1818.

\bibitem{Hu2021}
J.~Hu, P.~Bhowmick, I.~Jang, F.~Arvin, and A.~Lanzon, ``A decentralized cluster
  formation containment framework for multirobot systems,'' \emph{{IEEE}
  Transactions on Robotics}, vol.~37, no.~6, pp. 1936--1955, dec 2021.

\bibitem{Ferrante_2013}
E.~Ferrante, A.~E. Turgut, M.~Dorigo, and C.~Huepe, ``Elasticity-based
  mechanism for the collective motion of self-propelled particles with
  springlike interactions: A model system for natural and artificial swarms,''
  \emph{Physical Review Letters}, vol. 111, no.~26, p. 268302, dec 2013.

\bibitem{GomezNava2022}
L.~G{\'{o}}mez-Nava, R.~Bon, and F.~Peruani, ``Intermittent collective motion
  in sheep results from alternating the role of leader and follower,''
  \emph{Nature Physics}, vol.~18, no.~12, pp. 1494--1501, oct 2022.

\bibitem{Nguyen2020}
H.~Nguyen, T.~Dang, and K.~Alexis, ``The reconfigurable aerial robotic chain:
  Modeling and control,'' in \emph{2020 {IEEE} International Conference on
  Robotics and Automation ({ICRA})}.\hskip 1em plus 0.5em minus 0.4em\relax
  {IEEE}, may 2020.

\bibitem{Drupt2022}
J.~Drupt, C.~Dune, A.~I. Comport, S.~Seillier, and V.~Hugel,
  ``Inertial-measurement-based catenary shape estimation of underwater cables
  for tethered robots,'' in \emph{2022 {IEEE}/{RSJ} International Conference on
  Intelligent Robots and Systems ({IROS})}.\hskip 1em plus 0.5em minus
  0.4em\relax {IEEE}, oct 2022.

\bibitem{Laranjeira2020}
M.~Laranjeira, C.~Dune, and V.~Hugel, ``Catenary-based visual servoing for
  tether shape control between underwater vehicles,'' \emph{Ocean Engineering},
  vol. 200, p. 107018, mar 2020.

\bibitem{He2019}
S.~He, H.-S. Shin, and A.~Tsourdos, ``Trajectory optimization for target
  localization with bearing-only measurement,'' \emph{{IEEE} Transactions on
  Robotics}, vol.~35, no.~3, pp. 653--668, jun 2019.

\bibitem{Li2023}
J.~Li, Z.~Ning, S.~He, C.-H. Lee, and S.~Zhao, ``Three-dimensional bearing-only
  target following via observability-enhanced helical guidance,'' \emph{{IEEE}
  Transactions on Robotics}, vol.~39, no.~2, pp. 1509--1526, apr 2023.

\bibitem{Liang2018}
X.~Liang, H.~Wang, Y.-H. Liu, W.~Chen, and T.~Liu, ``Formation control of
  nonholonomic mobile robots without position and velocity measurements,''
  \emph{{IEEE} Transactions on Robotics}, vol.~34, no.~2, pp. 434--446, Apr.
  2018.

\bibitem{Soares_2013}
J.~M. Soares, A.~P. Aguiar, A.~M. Pascoal, and A.~Martinoli, ``Joint
  {ASV}/{AUV} range-based formation control: Theory and experimental results,''
  in \emph{2013 {IEEE} International Conference on Robotics and
  Automation}.\hskip 1em plus 0.5em minus 0.4em\relax {IEEE}, may 2013.

\bibitem{Berlinger2021}
F.~Berlinger, M.~Gauci, and R.~Nagpal, ``Implicit coordination for 3d
  underwater collective behaviors in a fish-inspired robot swarm,''
  \emph{Science Robotics}, vol.~6, no.~50, jan 2021.

\bibitem{Kottege2010}
N.~Kottege and U.~R. Zimmer, ``Underwater acoustic localization for small
  submersibles,'' \emph{Journal of Field Robotics}, vol.~28, no.~1, pp. 40--69,
  nov 2010.

\bibitem{Krupinski_2017}
S.~Krupinski, G.~Allibert, M.-D. Hua, and T.~Hamel, ``An inertial-aided
  homography-based visual servo control approach for (almost) fully actuated
  autonomous underwater vehicles,'' \emph{{IEEE} Transactions on Robotics},
  vol.~33, no.~5, pp. 1041--1060, Oct. 2017.

\bibitem{Chaumette2004}
F.~Chaumette, ``Image moments: A general and useful set of features for visual
  servoing,'' \emph{{IEEE} Transactions on Robotics}, vol.~20, no.~4, pp.
  713--723, Aug. 2004.

\bibitem{Liu2019}
X.~Liu, S.~S. Ge, and C.-H. Goh, ``Vision-based leader{\textendash}follower
  formation control of multiagents with visibility constraints,'' \emph{{IEEE}
  Transactions on Control Systems Technology}, vol.~27, no.~3, pp. 1326--1333,
  May 2019.

\bibitem{5299228}
G.~L. Mariottini, F.~Morbidi, D.~Prattichizzo, N.~Vander~Valk, N.~Michael,
  G.~Pappas, and K.~Daniilidis, ``Vision-based localization for
  leader–follower formation control,'' \emph{IEEE Transactions on Robotics},
  vol.~25, no.~6, pp. 1431--1438, 2009.

\bibitem{Lai_2022}
N.~Lai, Y.~Chen, J.~Liang, B.~He, H.~Zhong, H.~Zhang, and Y.~Wang, ``Image
  dynamics-based visual servo control for unmanned aerial manipulatorl with a
  virtual camera,'' \emph{{IEEE}/{ASME} Transactions on Mechatronics}, vol.~27,
  no.~6, pp. 5264--5274, dec 2022.

\bibitem{Li2021c}
J.~Li, H.~Xie, K.~H. Low, J.~Yong, and B.~Li, ``Image-based visual servoing of
  rotorcrafts to planar visual targets of arbitrary orientation,'' \emph{{IEEE}
  Robotics and Automation Letters}, vol.~6, no.~4, pp. 7861--7868, oct 2021.

\bibitem{groves2021}
M.~Dimitrov, K.~Groves, D.~Howard, and B.~Lennox, ``Model identification of a
  small fully-actuated aquatic surface vehicle using a long short-term memory
  neural network,'' in \emph{2021 IEEE International Conference on Robotics and
  Automation (ICRA)}, 2021, pp. 5966--5972.

\bibitem{Groves2019}
K.~Groves, A.~West, K.~Gornicki, S.~Watson, J.~Carrasco, and B.~Lennox,
  ``{MallARD}: An autonomous aquatic surface vehicle for inspection and
  monitoring of wet nuclear storage facilities,'' \emph{Robotics}, vol.~8,
  no.~2, p.~47, Jun. 2019.

\bibitem{Fossen1995}
T.~I. Fossen and O.-E. Fjellstad, ``Nonlinear modelling of marine vehicles in 6
  degrees of freedom,'' \emph{Mathematical Modelling of Systems}, vol.~1,
  no.~1, pp. 17--27, jan 1995.

\bibitem{Wang2016}
J.~Wang and E.~Olson, ``{AprilTag 2: Efficient and robust fiducial
  detection},'' in \emph{2016 IEEE/RSJ International Conference on Intelligent
  Robots and Systems (IROS)}.\hskip 1em plus 0.5em minus 0.4em\relax IEEE, oct
  2016, pp. 4193--4198.

\bibitem{Meier2015}
L.~Meier, D.~Honegger, and M.~Pollefeys, ``{PX}4: A node-based multithreaded
  open source robotics framework for deeply embedded platforms,'' in \emph{2015
  {IEEE} International Conference on Robotics and Automation ({ICRA})}.\hskip
  1em plus 0.5em minus 0.4em\relax {IEEE}, may 2015.

\end{thebibliography}


\end{document}